%% file: acl_latex.tex
\documentclass[11pt]{article}

\usepackage[preprint]{acl}

\usepackage{times}
\usepackage{latexsym}

\usepackage[T1]{fontenc}

\usepackage[utf8]{inputenc}

\usepackage{microtype}

\usepackage{inconsolata}

\usepackage{graphicx}

\usepackage{enumitem}
\usepackage{algorithm}
\usepackage{algorithmic}
\usepackage{tcolorbox}
\usepackage{subcaption} 
\usepackage{amsmath}
\usepackage{amssymb}
\usepackage{booktabs} 
\usepackage{multirow}
\usepackage{dsfont}
\usepackage{tabularx} 
\newcolumntype{Y}{>{\centering\arraybackslash}X} 
\usepackage{makecell}
\usepackage{xcolor}
\usepackage{pifont} 
\tcbuselibrary{listings, skins, breakable} 

\title{GroupTravelBench: Benchmarking LLM Agents on Multi-User, Multi-Turn Travel Planning}

\author{
Xiang Cheng\textsuperscript{1,2},
Yulan Hu\textsuperscript{2}\thanks{Corresponding authors.},
Lulu Zheng\textsuperscript{2},\\
\textbf{Xiangwen Zhang\textsuperscript{2}},
\textbf{Zheng Pan\textsuperscript{2}},
\textbf{Xin Li\textsuperscript{2}},
\textbf{Yong Liu\textsuperscript{1}\footnotemark[1]} \\
\textsuperscript{1}Gaoling School of Artificial Intelligence, Renmin University of China \\
\textsuperscript{2}AMAP, Alibaba Group \\
\texttt{\{chengxiang1, liuyonggsai\}@ruc.edu.cn} \\
\texttt{\{huyulan, zll522441, zhangxiangwen.zxw, panzheng.pan, xin.li\}@alibaba-inc.com}
}

\begin{document}
\maketitle
\begin{abstract}

Travel planning in the real world is overwhelmingly a \textit{group} activity, yet existing LLM travel-planning benchmarks reduce it to a single user, where the field is approaching saturation. This single-user assumption sidesteps what makes group planning hard for an agent: discovering private preferences across multiple users, surfacing conflicts, and balancing utility against fairness. To bring the task back to its multi-user reality, we introduce \textbf{\textit{GroupTravelBench}}, the first benchmark for \textbf{multi-user, multi-turn} travel planning. Built from real user profiles, POI data, and ticket prices, it comprises 650 tasks across three difficulty levels, each running in a synchronous group-chat sandbox with cached tool data for reproducible offline evaluation. Beyond the multi-step reasoning and tool use that single-user benchmarks already test, GroupTravelBench probes three group-specific capabilities: \textit{(i) elicitation} of private preferences through multi-turn dialogue; \textit{(ii) coordination} of inter-user conflicts via compromise or subgrouping; and \textit{(iii) planning} that balances group utility against fairness. We pair this with a complementary evaluation framework combining rule-based outcome metrics and LLM-judge process metrics. Across a wide range of frontier models, even the strongest agents fall short on all four rule-based outcome metrics, with plan validity below 12\%, suggesting that group-level outcome quality is a key open challenge for LLM travel-planning\footnote{Code and data are publicly available at \url{https://github.com/small-xiangcheng/GroupTravelBench}}.

\end{abstract}
\input{introduction.tex}
\input{related_work.tex}

\input{benchmark.tex}
\input{experiments.tex}

\input{conclusion.tex}
\input{limitation.tex}


\bibliography{custom}

\appendix
\input{appendix.tex}

\end{document}

%% file: introduction.tex
\section{Introduction}
\label{sec:intro}

Travel planning in the real world is overwhelmingly a \emph{group} activity. Families, friends, and colleagues plan trips together, and their preferences are private, often partial, and frequently in conflict---where to go, how much to spend, when to wake up. Producing a satisfying itinerary therefore demands more than search and tool use: an agent must \emph{elicit} the preferences each user holds privately, \emph{coordinate} disagreements when they surface, and \emph{plan} an itinerary that balances overall utility with fairness. These three capabilities are routine in human group coordination, yet they sit outside the scope of how travel planning is currently studied for LLM agents.

A growing body of work has framed travel planning as a benchmark for LLM agents, starting from TravelPlanner~\cite{xie2024travelplanner} and extending toward stricter constraint checking~\cite{shao2026chinatravel}, finer scoring~\cite{qu2025tripscore}, multi-turn interaction~\cite{qin2025compass,cheng2025travelbench}, and dynamic disturbances~\cite{karmakar2025triptide}. Yet all of these reduce travel planning to a \emph{single} user, and on this surface the field is approaching saturation: TravelPlanner has been largely solved by symbolic methods, with reported pass rates above 97\%~\cite{solve-travelplanner}, and the strongest models on the most recent multi-turn travel-planning benchmark already approach 80\%~\cite{cheng2025travelbench}. Group-side work, in contrast, has long existed in operations research~\cite{sylejmani2017planning,jouyandeh2023personalized,liu2025optimizing}, where group planning is formulated as a combinatorial optimization problem and preferences are treated as fixed numerical vectors given upfront---skipping precisely the interactive parts that make group planning hard for LLM agents.

Bringing the planning task back to its multi-user reality changes the problem along two compounding axes. \textbf{(a)} The plan space grows combinatorially with group size, since every additional user introduces preference dimensions, schedule constraints, and conflict points that must be jointly satisfied---this stresses the underlying planning capability beyond what current single-user benchmarks can probe. \textbf{(b)} Multi-user planning demands an interactive loop of \emph{elicitation}, \emph{coordination}, and group-aware \emph{planning} that no single-user setting can test by construction, regardless of how complex its constraints. A benchmark targeting this regime therefore evaluates \emph{harder} versions of what existing benchmarks cover, while opening up a class of capabilities they cannot reach at all.

To operationalize this, we introduce \textbf{\textit{GroupTravelBench}}: 650 multi-user travel-planning tasks built from real user profiles, real POI data, and real-world ticket prices, organized into three difficulty levels. Each user's preferences follow a four-tier hierarchy (\textit{must} / \textit{prefer} / \textit{avoid} / \textit{reject}) and are tracked in three coexisting tables---original, agent-inferred, and post-compromise---so that elicitation and conflict resolution can be scored deterministically. Each task runs in a synchronous group-chat sandbox where one agent interacts with multiple user simulators, supported by 10 travel tools and over 250K cached real-world tool records for stable, reproducible offline evaluation.

We pair the benchmark with a comprehensive evaluation framework that combines two complementary views: rule-based metrics for outcome quality (preference completeness, group utility, group fairness, plan validity) and an LLM judge for process quality (hallucination, tool use, interaction, conflict coordination, plan humanization). Across a wide range of frontier models, even the strongest agents fall short across all four rule-based outcome metrics---plan validity stays below 12\% across all models---indicating that group-level outcome quality remains an important bottleneck.

\noindent\textbf{Main contributions.}\par
\noindent(1) \textbf{\textit{GroupTravelBench}}, the first benchmark for \textbf{multi-user, multi-turn} travel planning, jointly evaluating preference \emph{elicitation}, conflict \emph{coordination}, and group-aware \emph{planning}.\par
\noindent(2) A \textbf{reproducible interactive framework} that simulates real group-chat dynamics---@-mention scheduling, convergence checkpoints, and compromise or subgrouping---backed by a sandbox of 10 travel tools and over 250K cached real-world records for stable offline evaluation.\par
\noindent(3) An evaluation methodology that uses three preference tables to \textbf{deterministically score preference elicitation and conflict resolution}, complemented by an LLM judge for process quality.

%% file: related_work.tex
\section{Related Work}
\label{sec:related}

\subsection{Group Travel Planning}

Planning itineraries for groups with heterogeneous preferences has long been studied as combinatorial optimization in the operations research community. Early efforts extend the Tourist Trip Design Problem to multiple travelers by modeling individual preference profiles and inter-member relationships, and solve the resulting problem with metaheuristics such as tabu search and evolutionary algorithms~\cite{sylejmani2017planning,jouyandeh2023personalized}. Subsequent work enriches this formulation with sustainability objectives and fuzzy-set preference modeling~\cite{ruiz2022grasp} and with dynamic subgrouping or ``joining-and-forking'' strategies that let members temporarily separate when preferences diverge~\cite{liao2022time,alatiyyah2025novel}; more recent efforts cast the problem as a crowd-aware Markov decision process~\cite{liu2025optimizing} or as multi-party multi-objective optimization with bargaining game theory~\cite{kogoya2026automatic}.

Despite steady progress, these approaches encode preferences as pre-defined numerical vectors and restrict planning scope to POI sequencing. None features a conversational agent that actively elicits, clarifies, and reconciles conflicting requirements through dialogue,which limits their applicability in real-world settings.

\subsection{LLM-Based Single-User Travel Planning}

Building on TravelPlanner~\cite{xie2024travelplanner}, which first framed travel planning as tool-grounded itinerary construction under hard and commonsense constraints, this line of work has evolved along four axes. The first pushes \emph{constraint complexity}: ChinaTravel~\cite{shao2026chinatravel} adds DSL-based compositional constraints, while more recent benchmarks tighten cross-action coupling~\cite{wang2026worldtravel} and demand long-horizon global feasibility~\cite{zhang2026deepplanning,yang2026widehorizon}; on the solution side, solver-based methods~\cite{solve-travelplanner,shao-etal-personal_travel_solver} and decoupled or RL-trained planners~\cite{yuan2026decoupled,wang2026tourplanner} largely close the feasibility gap on these benchmarks. The second moves \emph{beyond feasibility}: TP-RAG~\cite{TP-RAG}, TripTailor~\cite{wang-etal-2025-triptailor}, and TripScore~\cite{qu2025tripscore} shift evaluation toward spatiotemporal rationality, personalization, and unified plan quality. The third models \emph{realistic interaction} with implicit or under-specified intent, where the agent must proactively clarify~\cite{zhang-ask-before-plan}, optimize user preferences across multi-turn dialogue~\cite{qin2025compass,deng2025retail}, or jointly elicit implicit preferences and flag unsolvable requests~\cite{cheng2025travelbench}. The fourth probes \emph{robustness and root causes}: TripTide~\cite{karmakar2025triptide} stress-tests itineraries under real-world disruptions, and a recent diagnostic study attributes residual failures to weak inference of implicit constraints and ineffective self-correction~\cite{zhang2026revisiting}.

Across this body of work, however, the unit of evaluation remains \emph{one} traveler's plan. No existing benchmark requires an agent to elicit preferences from multiple participants, surface inter-user conflicts, or balance group-level fairness---yet real travel planning is overwhelmingly a group activity, the gap our benchmark is designed to fill.

\subsection{LLM Agents and Agentic RL in Travel}

Building on ReAct-style tool use~\cite{yao2022react}, a growing body of work trains LLM agents with reinforcement learning along three threads. The first establishes general recipes for tool-use RL, showing that outcome-based or hierarchically structured rewards suffice for learning strategic, multi-tool reasoning~\cite{qian2025toolrl,feng2025retool,dong2025tool-star}. The second targets practical bottlenecks of agentic RL---scaling to long-horizon trajectories~\cite{wu2026demystifying} and stabilizing reward signals on open-ended tasks via tournament-based relative ranking~\cite{zhang2026arenarl}. The third applies agentic RL directly to travel: DeepTravel~\cite{ning2025deeptravel} introduces hierarchical reward modeling and failure replay buffers, and TourPlanner~\cite{wang2026tourplanner} adds constraint-gated optimization with multi-path consensus. All such advances, however, target \emph{single-user, single-objective} settings: no current training framework addresses multi-user planning, where the agent must elicit preferences from several participants, negotiate conflicts, and optimize for group fairness. Our benchmark exposes precisely this capability gap as a target for future training.

%% file: benchmark.tex

\section{GroupTravelBench}
\label{sec:benchmark}

\begin{figure*}[t]
\centering
\includegraphics[width=\textwidth]{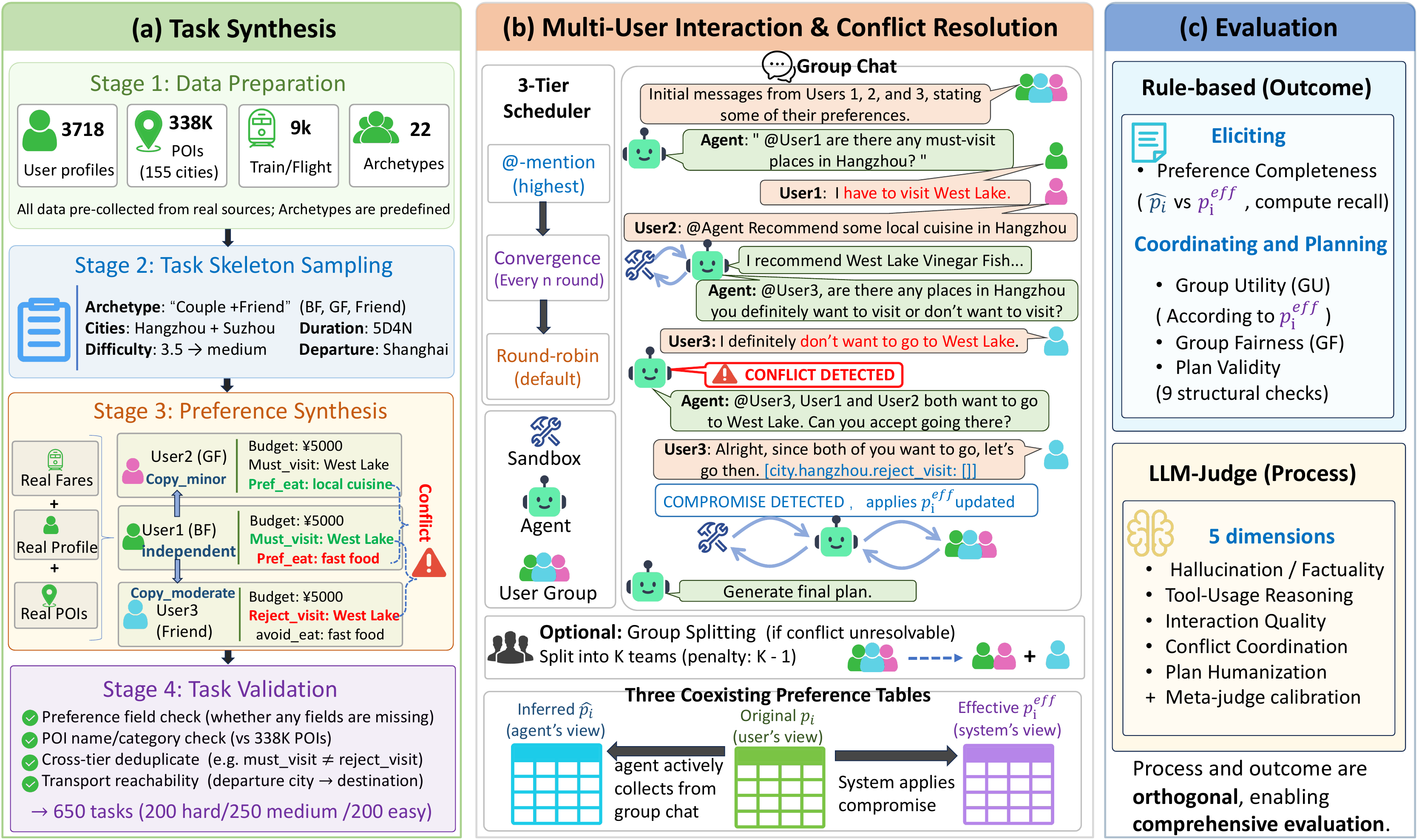}
\caption{Overview of \textit{GroupTravelBench}. The benchmark consists of three tightly coupled components: (a)~a task synthesis pipeline grounded in real-world data, (b)~a multi-user interaction framework that simulates group-chat planning, and (c)~an evaluation protocol that measures both final outcomes and interaction processes.}
\label{fig:overview}
\vspace{-8pt}
\end{figure*}

Figure~\ref{fig:overview} gives an overview of \textit{GroupTravelBench}. Preference modeling is foundational to \textit{GroupTravelBench}, as it directly shapes task synthesis, user simulator behavior, and how the evaluation protocol scores agent outputs. We therefore present this scheme first (§\ref{sec:preference}), and then describe task synthesis (§\ref{sec:synthesis}), multi-user interaction (§\ref{sec:interaction}), and the evaluation protocol (§\ref{sec:eval}).
\subsection{Preference Modeling}
\label{sec:preference}

A key challenge in group travel planning is how to model user preferences and evaluate conflict resolution in a deterministic manner. We argue that real-world preferences are inherently hierarchical: users may have options they strongly desire, clearly dislike, or feel neutral about. Such structured preferences naturally test an agent's ability to discover conflicts and resolve them effectively. A strong agent should be able to infer preferences from natural language, recognize their intensity, and use this information to mediate conflicts and generate a feasible itinerary.

To enable rule-based evaluation of both preference elicitation and conflict resolution, we design a structured preference table to model user preferences based on realistic travel needs (see Appendix~\ref{app:preference} for details). Each preference is associated with one of four intensity levels: \texttt{must}, \texttt{prefer}, \texttt{avoid}, and \texttt{reject}. The final itinerary receives different rewards or penalties depending on whether these preferences are satisfied or violated. For example, satisfying a \texttt{must} preference yields +2 points, whereas violating an \texttt{avoid} preference incurs a penalty of 1 point. During interaction, users express their preferences according to this table, and accepted compromises are reflected through updates to the preference table.

Three pieces of information must coexist for evaluation to be deterministic: what each user \emph{actually} prefers, what the agent \emph{believes} they prefer, and what they \emph{agree to} after compromise. We therefore maintain three coexisting preference tables for each user $i$: the \textbf{original preference table} $\mathbf{p}_i$, which contains the user's hidden ground-truth preferences; the \textbf{agent-inferred table} $\hat{\mathbf{p}}_i$, which records the agent's current belief about that user; and the \textbf{effective preference table} $\mathbf{p}_i^{\text{eff}}$, which applies all accepted compromises to the original preferences during natural-language interaction. When the agent identifies a conflict, it may explicitly ask a user to compromise. If the user is willing to compromise, the simulator responds in natural language and additionally emits a system-readable update. This update is validated by the framework and written into $\mathbf{p}_i^{\text{eff}}$. The update itself is not exposed to the agent. Instead, the agent must infer from the user's natural-language response whether a compromise has occurred and update its inferred preference table $\hat{\mathbf{p}}_i$ accordingly. This design allows us to evaluate not only conflict resolution, but also the agent's ability to recognize and track preference changes during interaction.

 


\subsection{Task Synthesis}
\label{sec:synthesis}

Real multi-user travel-planning data is hard to obtain at scale---it lives in private group chats and mixes implicit preferences with verbal hedging. Yet the underlying task template is highly regular: a fixed group of $N$ people heading to one or more destinations over a fixed number of days. This regularity lets us synthesize a large, diverse benchmark \emph{around} a real-world data backbone (POIs, ticket prices, user profiles), rather than fabricating content from scratch. We instantiate this idea via a four-stage pipeline---\emph{preparing} real data, \emph{sampling} task skeletons, \emph{generating} user preferences, and \emph{post-validating} the final tasks---with each stage controllable and grounded in real data, thereby balancing realism and distributional diversity.

\textbf{Stage 1: Data preparation.}
We first prepare four types of resources for task construction, including three types of real-world data and one manually defined archetype set. \textbf{POI data} contain about 338K entries collected from Chinese map services, including 100K attractions, 119K restaurants, and 119K hotels across 155 predefined cities, each annotated with categories and opening hours. \textbf{Transportation data} include pre-collected train and flight prices between city pairs. \textbf{User profiles} consist of 3,718 anonymized real user preferences covering demographics such as age, gender, marital status, and parenting status. \textbf{Group archetypes} cover 22 common travel settings for groups of size 2--6 (Appendix Table~\ref{tab:archetypes}).

\textbf{Stage 2: Task skeleton sampling.}
We then sample a task skeleton by jointly selecting a group archetype, a destination set (single city or 2--3 geographically adjacent cities), a departure city, a trip duration of 2--7 days, and a compromise configuration that specifies which users can make compromises. Sampling is popularity-weighted to reflect realistic travel demand. Difficulty labels are then assigned based on group size, trip length, and number of destinations, yielding 200 easy, 250 medium, and 200 hard tasks (Appendix~\ref{app:difficulty}).

\textbf{Stage 3: Preference table generation.}
Real groups share some tastes but diverge on others; generating preferences fully independently erases this correlation, while copying them across users erases individual diversity. We therefore generate each group member's hierarchical preference table with controllable inheritance. We first sample a role-consistent user profile from the profile pool; for example, a ``grandfather'' role is sampled from profiles with matching age and family attributes. We then provide the LLM with real city-specific POI candidates and transportation prices as context, and ask it to generate $user_i$'s original structured preference table $\mathbf{p_i}$ grounded in those inputs.

Preference tables include both \emph{global constraints}, such as per-person budget and transportation preferences, and \emph{city-level preferences}, such as attraction and food choices. To create realistic within-group diversity, we use three generation strategies based on the strength of user relationships: \textbf{copy\_minor}, where broad preferences are inherited but specific POI-level choices are regenerated; \textbf{copy\_moderate}, where only \emph{global constraints} are inherited while city-level preferences are regenerated; and \textbf{independent}, where the entire table is generated from scratch. These strategies allow group members to share some tastes without becoming unrealistically identical (Appendix~\ref{app:pref_gen}). We also generate one opening utterance per user, revealing only 1--3 preferences as conversational seeds, with wording aligned to preference strength. The model used for preference generation is Gemini-3-Flash~\cite{google_gemini3_2025}.

\textbf{Stage 4: Post-validation.}
LLM-based synthesis can hallucinate POIs or introduce internal inconsistencies, so we apply three validation checks---POI existence, cross-tier consistency, and transportation reachability (Appendix~\ref{app:pref_gen})---before releasing each task. Invalid generations trigger regeneration; tasks that pass are realistic, internally consistent, and solvable in principle.

\subsection{Multi-Party Interaction Framework}
\label{sec:interaction}
A central goal of \textit{GroupTravelBench} is to faithfully reproduce how multi-user travel planning unfolds in real chat groups. In practice, users @-mention the agent to ask for information, the agent @-mentions individual users to elicit private preferences or request compromises, and agent has to keep nudging the discussion toward consensus. To capture this loop within an evaluable framework, each task runs in a synchronous group-chat environment where one agent interacts with $N$ LLM-based user simulators through a shared ordered conversation history.

\textbf{Interaction protocol.}
The interaction begins with a broadcast of the pre-generated opening utterances and proceeds as a free-form dialogue under a simple scheduler: when the agent @-mentions a user, that user gains immediate speaking priority; otherwise, turns follow round-robin order. Every $\kappa$ rounds the agent is required to summarize the preferences and conflicts collected so far, providing periodic convergence signals. When two preferences cannot be reconciled, the agent may either request a compromise from a specific user or split the group into subgroups with parallel sub-itineraries; subgrouping makes planning easier and therefore incurs a penalty, treating it as a costly meta-decision rather than a free escape from coordination. Tool calls are routed through a sandbox of 10 travel tools backed by over 250K cached real-world records (Appendix~\ref{app:tools}), with retrieval-based fallback on cache misses for stable offline evaluation~\cite{cheng2025travelbench,guo2024stabletoolbench}. Per-difficulty round budgets, compromise and subgrouping mechanics, termination conditions, and more details are in Appendix~\ref{app:interaction_details}.

\textbf{User behavior constraints.}
To ensure the benchmark measures \emph{the agent's} elicitation ability rather than user-side cooperativeness, simulators are forbidden from proactively revealing preferences; they speak only when @-mentioned or when their strong preferences are violated. The full behavior contract and tone-mapping rules of users are in Appendix~\ref{app:user_sim}.

\subsection{Dataset Statistics}
\label{sec:stats}

Appendix~\ref{app:statistics} includes the main statistics of \textit{GroupTravelBench}. The benchmark contains 650 tasks (200 easy, 250 medium, and 200 hard), covering 2,748 preference-bearing users in total and an average of 4.2 such users per task. Pre-school children (under 8) are listed as participants in the group but carry no preference table, and are therefore excluded from this count. It spans 22 group archetypes, 143 destination cities, and 32 departure cities. In total, the benchmark includes nearly 63K atomic preference items grounded in 3,718 real user profiles and approximately 338K real POIs. Conflict is common by construction: 77.4\% of tasks contain at least one within-group preference conflict and the per-task mean is 7.52 (Appendix~\ref{app:pref_dist}), providing substantial coverage for evaluating conflict discovery and coordination.

\section{Evaluation Protocol}
\label{sec:eval}

We evaluate each run from two complementary perspectives: \textbf{outcome quality} and \textbf{process quality}. Rule-based metrics can verify the final itinerary against explicit constraints, but they cannot capture process quality or soft requirements such as accommodation for elderly travelers or children. We therefore additionally introduce an LLM judge~\cite{hashemi2024llm-rubric} to evaluate process quality. The two views are complementary rather than redundant: the LLM judge is explicitly instructed not to score aspects already covered by deterministic checks, such as formatting, time consistency, or exact numerical matching.

\subsection{Rule-Based Outcome Metrics}

\textbf{Preference Completeness (PC).}
PC measures how accurately the agent collects user preferences during dialogue. We compare the agent's final inferred table $\hat{\mathbf{p}}_i$ against the effective preference table $\mathbf{p}_i^{\text{eff}}$ of user $i$. For list-valued fields, we count matched items; for scalar fields, we require exact match. We report recall as the main metric:
\[
\text{PC} = 100 \cdot \frac{\sum_i \text{collected}_i}{\sum_i \text{possible}_i},
\]
where $i$ indexes users in the group.

\textbf{Group Utility (GU).}
GU measures how well the final itinerary satisfies group preferences after compromise. For each user $i$, we compute a utility score $u_i$ against $\mathbf{p}_i^{\text{eff}}$, where strong satisfaction and violation receive $\pm 2$, and weak satisfaction and violation receive $\pm 1$. The score covers attractions, food, transportation mode, hotel type, activity intensity, and budget. If the agent uses subgrouping, we apply a split penalty: each split event $e$ that divides the group into $K_e$ subgroups incurs a penalty of $K_e - 1$. We report:
\[
\text{GU} = \frac{1}{N}\left(\sum_{i=1}^{N} u_i - \sum_e (K_e - 1)\right),
\]
where $N$ is the group size.

\textbf{Group Fairness (GF).}
A plan with high overall utility may still neglect some users. GF measures the balance of utility across users:
\[
\text{GF} = 100 \cdot \frac{\min_i u_i}{\max_i u_i},
\]

\textbf{Plan Validity (PV).}
PV measures whether the generated itinerary is physically feasible. We apply nine deterministic checks spanning transportation, lodging, scheduling, and cost completeness (full list in Appendix~\ref{app:pv_checks}). A plan is counted as valid only if it passes all checks:
\[
\text{PV} = \mathds{1}[\text{issues}(\hat{y}) = \emptyset],
\]
where $\hat{y}$ is the agent's final plan and $\text{issues}(\cdot)$ returns the list of failed checks.
\subsection{LLM-Based Process Evaluation}

Rule-based metrics capture only verifiable outcomes. We therefore employ an LLM judge to evaluate process quality along five dimensions: \textbf{hallucination}, \textbf{tool-use reasoning}, \textbf{interaction quality}, \textbf{conflict coordination}, and \textbf{plan humanization}. Each dimension is scored on a 1--5 scale with evidence grounded in specific dialogue turns. Detailed evaluation dimensions are provided in Appendix~\ref{app:llm_judge_dims}.

A single LLM judge can be biased or inconsistent across runs, so we add a \textbf{meta-evaluation} step that calibrates each judge verdict: a second LLM reviews the first judge's reasoning and produces a calibration score $s^{\text{meta}} \in \{1,\dots,5\}$ that scales the final score. Calibration setup and rater-agreement statistics appear in Appendix~\ref{app:meta_judge}. The final LLM-judge score for a task is:
\vspace{-5pt}
\[
S^{\text{LJ}} = 100 \cdot \frac{1}{5}\sum_{d=1}^{5}\frac{r_d-1}{4}\cdot \frac{s^{\text{meta}}}{5},
\]
\vspace{-5pt}
where $r_d$ is the score in the $d$-th dimension.

\subsection{Reporting}
We report all five metrics separately: PC, GU, GF, PV, and $S^{\text{LJ}}$. All five are first computed per task and then averaged over the evaluation set $\mathcal{D}$. The four rule-based outcome metrics align with the capabilities \textit{GroupTravelBench} targets---PC measures \emph{elicitation}, GU and GF measure \emph{coordination} into a group-aware plan, and PV measures the feasibility floor that all of the above must respect---while $S^{\text{LJ}}$ supplies process-level signals that no deterministic check can capture. Because GU cannot be normalized without access to the optimal achievable utility, we additionally report GU as a standalone metric and the average of the other four metrics.

%% file: experiments.tex

\section{Experiments}
\label{sec:experiments}

\begin{table*}[t]
\small
\centering
\setlength{\tabcolsep}{5pt}
\renewcommand{\arraystretch}{0.9}
\caption{Main results on \textit{GroupTravelBench} across models and difficulty levels. 
Judge denotes the average score assigned by the LLM-judge, and Other denotes the average of PC, GF, PV, and Judge. 
For Qwen3 variants, Th denotes the thinking mode and It denotes the instruction-following variant. GPT-5.1 is evaluated in instruct mode, whereas the other models are used with reasoning enabled. 
Best results in each column are shown in \textbf{bold}.}

\label{tab:main}
\begin{tabularx}{0.95\textwidth}{l *{5}{Y} *{6}{Y}}
\toprule
\multirow{2}{*}{\textbf{Model}} &
\multicolumn{5}{c}{\textbf{All (650)}} &
\multicolumn{2}{c}{\textbf{Easy (200)}} &
\multicolumn{2}{c}{\textbf{Medium (250)}} &
\multicolumn{2}{c}{\textbf{Hard (200)}} \\
\cmidrule(lr){2-6}\cmidrule(lr){7-8}\cmidrule(lr){9-10}\cmidrule(lr){11-12}
& \textbf{GU} & \textbf{PC} & \textbf{GF} & \textbf{PV} & \textbf{Judge}
& \textbf{GU} & \textbf{Other}
& \textbf{GU} & \textbf{Other}
& \textbf{GU} & \textbf{Other} \\
\midrule
\multicolumn{12}{l}{\textbf{Frontier Models}} \\
\midrule
DeepSeek-V4-Pro & \textbf{10.47} & \textbf{64.0} & \textbf{54.2} & 8.0 & \textbf{92.3} & \textbf{9.46} & \textbf{62.2} & \textbf{11.35} & \textbf{53.5} & \textbf{10.36} & \textbf{48.1} \\
GPT-5.1         & 7.21 & 31.0 & 48.1 & 1.0 & 85.2 & 6.35 & 46.8 & 7.44 & 40.8 & 7.78 & 36.5 \\
Qwen3.6-Max     & 9.66 & 31.0 & 52.1 & 7.0 & 57.3 & 8.54 & 43.4 & 10.03 & 36.7 & 10.33 & 31.5 \\
\midrule
\multicolumn{12}{l}{\textbf{Lightweight Models}} \\
\midrule
Qwen3.5-Plus    & 8.63 & 30.0 & 44.9 & \textbf{12.0} & 62.2 & 7.57 & 42.0 & 9.02 & 38.0 & 9.20 & 32.4 \\
Qwen3.5-Flash   & 6.97 & 21.0 & 41.9 & 1.0 & 43.7 & 6.20 & 31.4 & 7.05 & 26.7 & 7.65 & 23.3 \\
Qwen3-235B-Th   & 7.35 & 26.0 & 49.0 & 3.0 & 45.4 & 6.47 & 37.9 & 7.40 & 29.9 & 8.15 & 25.7 \\
Qwen3-30B-Th    & 6.20 & 18.0 & 48.6 & 1.0 & 21.3 & 5.27 & 27.6 & 6.26 & 21.4 & 7.07 & 18.4 \\
Qwen3-4B-Th     & 5.72 & 22.0 & 47.6 & 0.0 & 24.8 & 4.89 & 29.7 & 5.64 & 22.3 & 6.64 & 19.3 \\
Qwen3-4B-It     & 6.12 & 18.0 & 41.7 & 1.0 & 22.9 & 5.22 & 26.1 & 6.05 & 20.1 & 7.10 & 16.9 \\
\bottomrule
\end{tabularx}
\vspace{-5pt}
\end{table*}

\subsection{Experimental Setup}
\label{sec:setup}
\textbf{Models.}
We report results on nine LLMs spanning both frontier and lightweight models. The \textit{frontier models} include GPT-5.1~\cite{openai_gpt51_2025}, DeepSeek-V4-Pro~\cite{deepseekv4}, and Qwen3.6-Max~\cite{yang2025qwen3}, while the \textit{lightweight models} mainly consist of Qwen reasoning models with different parameter sizes. We find that some models, especially smaller instruct models, struggle to complete this task. A detailed discussion is provided in Appendix~\ref{app:excluded_models}.

\textbf{Other Details.}
Each model is evaluated on the full 650-task benchmark with $T=3$ independent trials per task, using an agent sampling temperature of 0.7. We report the mean score across trials. We use DeepSeek-V4-Flash as the user simulator, with temperature set to 0 to improve reproducibility. For LLM-as-judge evaluation, we use Gemini-3-Flash with meta-judge calibration. We adopt difficulty-aware interaction budgets by scaling the maximum number of interaction rounds and convergence intervals according to task difficulty: Easy (15 rounds, $\kappa=3$), Medium (20 rounds, $\kappa=4$), and Hard (25 rounds, $\kappa=5$). More details are provided in Appendix~\ref{app:exp_details}.

\subsection{Main Results}
\label{sec:main_results}

Table~\ref{tab:main} reports five evaluation dimensions on the full benchmark and per-difficulty breakdowns.

Several findings emerge from Table~\ref{tab:main}. 
(1)~DeepSeek-V4-Pro outperforms all other models on almost every dimension, especially on Preference Completeness (64.0\% vs.\ $\leq$31\% for all others), confirming that proactive multi-turn preference elicitation is crucial. (2)~Group Fairness remains below 55\% for \emph{all} models, revealing a shared inability to distribute utility equitably across group members. (3)~Plan Validity is alarmingly low ($\leq$12\%), indicating that most models generate structurally flawed itineraries. (4)~GPT-5.1 achieves the second-highest LLM-Judge score (85.2\%) while ranking only fifth on GU, suggesting strong interaction quality but weak final outcome optimization. (5)~The task is extremely challenging for smaller instruction-tuned models: most open-source instruction-tuned models we tested (Appendix~\ref{app:excluded_models}) fail to produce well-formed tool calls as the interaction sequence becomes longer, whereas GPT-5.1 succeeds despite also being an instruction-following model. This result suggests that sufficient model capacity is a prerequisite for reliable multi-turn agentic behavior in this setting.

\begin{figure}[t]
\centering
\includegraphics[width=\columnwidth]{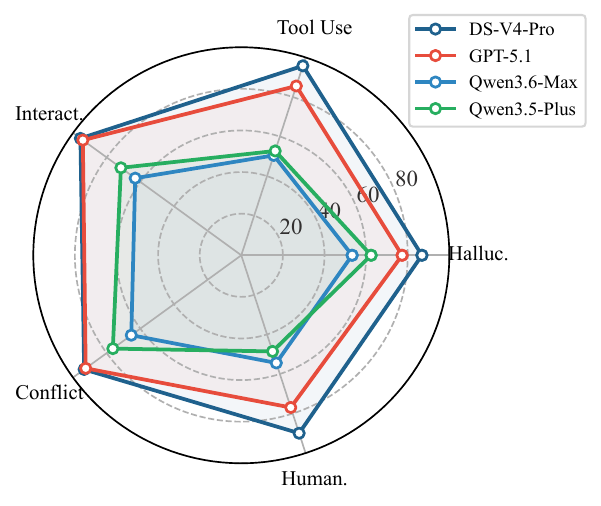}
\caption{LLM-Judge sub-dimension profiles. DeepSeek-V4-Pro leads uniformly; GPT-5.1 excels on interaction/conflict but lags on hallucination and humanization.}
\label{fig:radar}
\vspace{-10pt}
\end{figure}

\textbf{LLM-Judge sub-dimension profile.}
Figure~\ref{fig:radar} compares four models across the five process-quality dimensions. DS-V4-Pro leads uniformly; GPT-5.1 nearly matches it on Interaction and Conflict but falls behind on Hallucination (77.3 vs.\ 86.8) and Humanization (76.9 vs.\ 90.0). This breakdown clearly highlights where each model excels and where it remains limited.

\begin{table*}[t]
\small
\centering
\setlength{\tabcolsep}{5pt}
\renewcommand{\arraystretch}{0.9}
\caption{Offline (isolated sandbox) vs.\ Online (live API) comparison. \textit{Hit} is the cache hit rate and \textit{Tools} is the average number of tool calls. Score differences fall within inter-trial variance, confirming sandbox fidelity.}
\label{tab:offline_online}
\begin{tabularx}{0.95\textwidth}{l *{2}{Y} *{2}{Y} *{2}{Y} *{2}{Y}}
\toprule
\multirow{2}{*}{\textbf{Model}} &
\multicolumn{2}{c}{\textbf{GU}} &
\multicolumn{2}{c}{\textbf{Other (\%)}} &
\multicolumn{2}{c}{\textbf{Hit (\%)}} &
\multicolumn{2}{c}{\textbf{Tools}} \\
\cmidrule(lr){2-3}\cmidrule(lr){4-5}\cmidrule(lr){6-7}\cmidrule(lr){8-9}
& \textbf{Off.} & \textbf{On.}
& \textbf{Off.} & \textbf{On.}
& \textbf{Off.} & \textbf{On.}
& \textbf{Off.} & \textbf{On.} \\
\midrule
GPT-5.1         & 7.21 & 7.14 & 41.33 & 40.78 & 37.7 & 36.9 & 12.29 & 12.59 \\
DeepSeek-V4-Pro & 10.47 & 10.54 & 54.62 & 54.45 & 66.8 & 67.2 & 33.83 & 34.88 \\
Qwen3.5-Flash   & 6.97 & 6.92 & 26.88 & 27.11 & 62.6 & 61.5 & 20.32 & 20.29 \\
Qwen3.6-Max     & 9.66 & 9.69 & 36.84 & 37.26 & 61.7 & 62.3 & 20.00 & 19.93 \\
\bottomrule
\end{tabularx}
\vspace{-10pt}
\end{table*}

\subsection{Reliability}
\label{sec:reliability}

\paragraph{Cross-trial stability.}
Table~\ref{tab:stability} shows that our results are highly reproducible across three independent runs: the standard deviation is at most 0.17 for Group Utility and at most 0.30 for Other. We attribute the remaining variation primarily to stochasticity introduced by the agent sampling temperature (0.7). Overall, these small deviations indicate that both our inference pipeline and evaluation framework are stable and reproducible.

\begin{table}[t]
\small
\centering
\setlength{\tabcolsep}{4pt}
\renewcommand{\arraystretch}{0.95}
\caption{Cross-trial stability (mean $\pm$ std over $T=3$ independent runs). }
\label{tab:stability}
\begin{tabular}{lcc}
\toprule
\textbf{Model} & \textbf{GU} & \textbf{Other (\%)} \\
\midrule
GPT-5.1         & 7.20 $\pm$ 0.02 & 41.04 $\pm$ 0.21 \\
DeepSeek-V4-Pro & 10.47 $\pm$ 0.05 & 54.82 $\pm$ 0.30 \\
Qwen3.5-Flash   & 7.02 $\pm$ 0.05 & 27.01 $\pm$ 0.10 \\
Qwen3.5-Plus    & 8.76 $\pm$ 0.17 & 37.31 $\pm$ 0.22 \\
\bottomrule
\end{tabular}
\vspace{-10pt}
\end{table}


\paragraph{Offline/online consistency.}
\label{sec:offline_online}
Table~\ref{tab:offline_online} compares \texttt{ISOLATED} (cache + LLM simulator) and \texttt{ONLINE} (live API) modes across four models. Score differences are negligible: $|\Delta\text{GU}| \leq 0.07$ and $|\Delta\text{Other}| \leq 0.55$, well within inter-trial variance. Cache hit rates and tool call counts also remain nearly identical between the two modes ($|\Delta\text{Hit}| \leq 1.1$ percentage points)---confirming that the embedding-retrieval + ICL simulation faithfully replicates real API behavior without materially affecting model decisions.

\begin{figure}[t]
\vspace{3pt}
\centering
\includegraphics[width=\columnwidth]{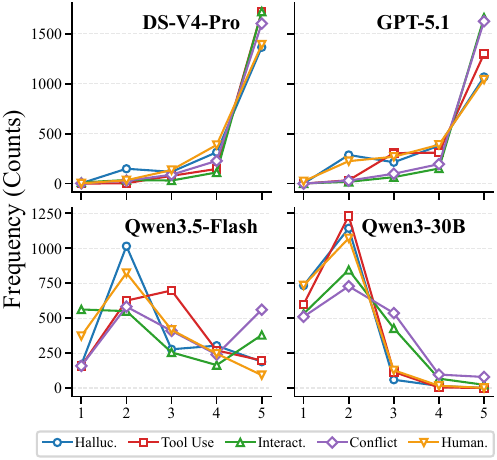}
\caption{LLM-Judge score distribution per dimension for four models. Top models cluster at 4--5; bottom models show dimension-dependent failure patterns.}
\label{fig:judge_dist}
\vspace{-10pt}
\end{figure}

\paragraph{LLM-Judge score distribution.}
Figure~\ref{fig:judge_dist} shows the per-dimension rating distribution for four models spanning the performance spectrum. Top-performing models are concentrated at scores 4--5, with DS-V4-Pro achieving score-5 rates above 70\% on Tool Use, Interaction, and Conflict. In contrast, lower-performing models show markedly different patterns: Qwen3.5-Flash peaks at score 2 on Hallucination, indicating pervasive fabrication, while Qwen3-30B clusters at scores 1--2 on most dimensions except Conflict, where a bimodal distribution suggests occasional but inconsistent coordination attempts. We further validate that the LLM-judge scores align with human preferences on 100 sampled trajectories (Appendix~\ref{app:judge_human_validation}).

%% file: conclusion.tex
\section{Conclusion}
We introduce \textbf{GroupTravelBench}, the first benchmark for multi-user, multi-turn travel planning with LLM agents. The benchmark synthesizes 650 tasks across 22 group archetypes and three difficulty levels, grounded in 3,718 real user profiles, ${\sim}$338K real POIs, and real-world transportation prices; each task runs in a group-chat sandbox backed by over 250K cached real-world tool records for stable offline evaluation. By coupling structured preference tables with this sandbox, GroupTravelBench evaluates three capabilities that single-user benchmarks cannot test: \emph{elicitation} of private preferences, \emph{coordination} of conflicts via compromise or subgrouping, and group-aware \emph{planning} that balances utility and fairness. We pair the benchmark with a complementary evaluation framework combining four rule-based outcome metrics with a meta-judge-calibrated LLM judge for process quality. Across a wide range of experiments, even the strongest agents fall short across all four rule-based outcome metrics---plan validity remains below 12\% across all models---indicating that robust multi-party travel planning remains far from solved and pointing to a clear training and evaluation surface for future LLM agents.

%% file: limitation.tex
\section*{Limitations}

\paragraph{Group Utility normalization.}
While our rule-based metrics deterministically score preference elicitation and conflict resolution, the search space of feasible group itineraries is too large to admit a tractable optimum for Group Utility (GU). As a result, GU cannot be normalized to a $[0,100]$ scale, and we cannot report a single overall average across all five metrics.

\paragraph{Limited interaction setting.}
To simplify evaluation and avoid the confounding effects of excessive user involvement, we restrict the LLM to generating only one final plan. In more realistic settings, however, an LLM would iteratively refine the plan through ongoing discussion and interaction with users. Our current evaluation does not cover this interactive planning process, which may create a gap between our setup and real-world application scenarios. Future work could explore how to introduce such settings in a controlled manner, enabling evaluation of LLM capabilities under more realistic conditions.

\paragraph{Geographic and cultural scope.}
All tasks are grounded in Chinese domestic travel, using Chinese map services, POI data, and transportation APIs. While the benchmark design is general, extending it to other countries or cross-border travel would require new data sources, currency handling, and visa-related constraints. Findings about model performance may not directly transfer to non-Chinese travel contexts.

%% file: appendix.tex
\clearpage

\section{Benchmark Details}
\label{app:benchmark}

This appendix documents the components that define the benchmark itself: how user preferences are structured, how groups are organized, how user simulators behave during interaction, what dimensions the LLM judge scores, and how the released 650 tasks are distributed across groups, cities, time, and difficulty.

\subsection{Preference Schema}
\label{app:preference}

Each user's preference table $\mathbf{p}_i$ follows a hierarchical structure with two levels: \emph{global constraints} (trip-wide) and \emph{city-specific preferences} (per destination city). Table~\ref{tab:preference_schema} provides the complete field inventory.

\begin{table*}[t]
\centering
\caption{Complete preference schema. \textit{Tier} indicates the strength level used in Group Utility scoring. Strong tiers (must/reject) carry $\pm2$ weight; weak tiers (prefer/avoid) carry $\pm1$ weight.}
\label{tab:preference_schema}
\small
\setlength{\tabcolsep}{5pt}
\renewcommand{\arraystretch}{1.15}
\begin{tabular}{@{}l l l c l@{}}
\toprule
\textbf{Level} & \textbf{Field} & \textbf{Description} & \textbf{Type} & \textbf{Tier} \\
\midrule
\multirow{9}{*}{Global}
& \texttt{avg\_budget} & Per-person budget cap (CNY) & scalar & strong ($-2$) \\
& \texttt{transport.must} & Transport modes the user insists on & list & strong ($+2$) \\
& \texttt{transport.reject} & Transport modes the user refuses & list & strong ($-2$) \\
& \texttt{transport.prefer} & Transport modes mildly preferred & list & weak ($+1$) \\
& \texttt{transport.avoid} & Transport modes mildly avoided & list & weak ($-1$) \\
& \texttt{intensity.max\_poi\_per\_day} & Max POIs per day before penalty & scalar & strong ($-2$) \\
& \texttt{intensity.max\_active\_hours} & Max active hours per day & scalar & strong ($-2$) \\
& \texttt{hotel\_preference.prefer} & Hotel categories mildly preferred & list & weak ($+1$) \\
& \texttt{hotel\_preference.avoid} & Hotel categories mildly avoided & list & weak ($-1$) \\
\midrule
\multirow{8}{*}{\shortstack[l]{City-\\specific}}
& \texttt{attractions.must\_visit} & POI names the user must visit & list & strong ($+2$) \\
& \texttt{attractions.reject\_visit} & POI names the user refuses to visit & list & strong ($-2$) \\
& \texttt{attractions.category\_pref.positive} & Attraction categories preferred & list & weak ($+1$) \\
& \texttt{attractions.category\_pref.negative} & Attraction categories disliked & list & weak ($-1$) \\
& \texttt{food.must\_eat} & Foods/restaurants the user must try & list & strong ($+2$) \\
& \texttt{food.reject\_eat} & Foods/restaurants the user refuses & list & strong ($-2$) \\
& \texttt{food.prefer\_eat} & Foods the user mildly prefers & list & weak ($+1$) \\
& \texttt{food.avoid\_eat} & Foods the user mildly avoids & list & weak ($-1$) \\
\bottomrule
\end{tabular}
\end{table*}

\paragraph{Design rationale: specific instances vs.\ category-level preferences.}
The choice of \emph{specific instances vs.\ categories} per field reflects how people actually voice travel preferences. For \textbf{attractions} and \textbf{food}, strong tiers (\texttt{must\_visit} / \texttt{reject\_visit}, \texttt{must\_eat} / \texttt{reject\_eat}) take specific POI or dish names---users usually have a small set of places or foods they actively want or refuse (``I must see the Forbidden City''; ``I refuse to eat hotpot''). Weak tiers (\texttt{prefer} / \texttt{avoid}), in contrast, are stated as broader categories (e.g., \emph{museum} vs.\ \emph{theme park}; \emph{Cantonese} vs.\ \emph{fast food}), because casual preferences typically come at the category level. For \textbf{hotels}, all four tiers are category-level (\emph{economy} / \emph{comfort} / \emph{business} / \emph{luxury}) since users rarely name specific hotels in conversation. For \textbf{transportation}, all four tiers are restricted to a fixed mode set (flight, train, high-speed rail, self-driving). Finally, \textbf{budget} and \textbf{activity intensity} are scalar caps that bound the entire trip rather than enumerated lists. Together, these per-field choices let a single schema cover both \emph{deliberate} preferences (a specific must-see) and \emph{categorical} taste (a general dislike of certain food types), giving a comprehensive yet realistic picture of each user.

\paragraph{Strength tiers and user-simulator tone mapping.}
The four-tier system serves a dual purpose. In \emph{evaluation}, it determines the scoring weight: strong-tier items (must, reject, budget cap, intensity cap) carry $\pm2$ points, while weak-tier items (prefer, avoid, category preferences, hotel preferences) carry $\pm1$ points. The asymmetry between positive and negative tiers is deliberate: for positive tiers (must/prefer), satisfying the preference earns credit but \emph{not} satisfying it incurs no penalty; for negative tiers (reject/avoid), violating the preference incurs a penalty but \emph{not} violating it earns no credit. This reflects the real-world semantics that fulfilling a wish is a bonus, while violating a taboo is a failure.

In \emph{interaction}, the tier directly controls the user simulator's conversational tone. When expressing a \texttt{must} preference, the user speaks with a firm, non-negotiable tone (e.g., ``I must take a flight''); when expressing a \texttt{prefer}, the tone is mild and flexible (e.g., ``If possible, I would prefer to fly''). This creates a natural signal that the agent must learn to interpret: the strength of a preference is encoded in \emph{how} the user says it, not in an explicit label. The user simulator is strictly forbidden from using field names (e.g., \texttt{must\_visit}) in conversation.

\subsection{Group Archetypes}
\label{app:archetypes}

We curate 22 social archetypes covering group sizes $N \in \{2, \ldots, 6\}$. Table~\ref{tab:archetypes} lists each archetype with its group size, member roles, social tags, and number of tasks in the released benchmark. Each archetype is additionally annotated with:

\begin{itemize}[leftmargin=*,itemsep=2pt]
\item \textbf{Role slots}: named roles for each member (e.g., ``boyfriend'', ``grandmother'', ``roommate~A''), used to condition user-profile sampling and preference generation.
\item \textbf{Social dynamics description}: a free-text paragraph (in Chinese) describing the typical decision-making dynamics, common sources of conflict, and who is most likely to compromise. These descriptions guide the LLM during preference synthesis and initial-message generation.
\item \textbf{Compromise patterns}: an exhaustive enumeration of all plausible subsets of members who might agree to compromise in a given task. At synthesis time, one pattern is sampled per task to set each user's \texttt{compromisable} flag.
\end{itemize}

\begin{table*}[t]
\centering
\caption{Distribution of the 22 social archetypes across group size in the released 650-task benchmark. All role slots, tag annotations, and admissible compromise patterns are released alongside the dataset.}
\label{tab:archetypes}
\small
\setlength{\tabcolsep}{8pt}
\renewcommand{\arraystretch}{1.15}
\begin{tabular}{clllr}
\toprule
$N$ & \textbf{Archetype} & \textbf{Roles (abbreviated)} & \textbf{Tags} & \textbf{\#Tasks} \\
\midrule
\multirow{4}{*}{2}
 & Couple & M, F & romance & 17 \\
 & Married couple & husband, wife & ritual & 13 \\
 & Female friends & F-A, F-B & food, photo & 24 \\
 & Male friends & M-A, M-B & outdoor, adventure & 20 \\
\midrule
\multirow{6}{*}{3}
 & Couple + female friend & BF, GF, single-F & mixed & 21 \\
 & Couple + male friend & BF, GF, single-M & mixed & 18 \\
 & Three female friends & F-A, F-B, F-C & democratic & 25 \\
 & Three male friends & M-A, M-B, M-C & free-form & 23 \\
 & Family w/ toddler & father, mother, child (0--8) & parenting & 16 \\
 & Family w/ school-age & father, mother, child (8--18) & parenting & 18 \\
\midrule
\multirow{6}{*}{4}
 & Two couples & BF-A, GF-A, BF-B, GF-B & dual-CP & 21 \\
 & Couple + two friends & BF, GF, F-A, F-B & mixed & 17 \\
 & Four female friends & F-A--D & democratic & 28 \\
 & Four male friends & M-A--D & adventure & 19 \\
 & Family w/ two children & father, mother, teen, child & age-gap & 21 \\
 & Couple + parents & husband, wife, father-in-law, mother-in-law & two-gen & 20 \\
\midrule
\multirow{3}{*}{5}
 & Five female friends & F-A--E & large group & 37 \\
 & Five male friends & M-A--E & large group & 46 \\
 & Family + grandparents & father, mother, child, grandpa, grandma & three-gen & 41 \\
\midrule
\multirow{3}{*}{6}
 & Three-gen large family & grandpa, grandma, father, mother, teen, child & complex & 80 \\
 & College dorm & roommate A--F & familiar but diverse & 53 \\
 & Three couples & BF/GF $\times$ 3 & triple-CP & 72 \\
\bottomrule
\end{tabular}
\end{table*}

\subsection{Data Field Definitions}
\label{app:fields}

Each task in \texttt{test.jsonl} is a JSON object containing the fields listed in Table~\ref{tab:fields}.

\begin{table*}[t]
\caption{Field definitions for a \textit{GroupTravelBench} task instance.}
\label{tab:fields}
\centering
\small
\renewcommand{\arraystretch}{1.1}
\begin{tabularx}{0.95\textwidth}{@{} >{\raggedright\arraybackslash\texttt}p{0.18\textwidth} X @{}}
\toprule
\textbf{\textrm{Field}} & \textbf{Description} \\
\midrule
task\_id & A unique identifier for each task (e.g., \texttt{task\_004329}). \\
\midrule
query & A coarse natural-language travel request specifying destination cities, trip duration, and group size. This is the only information visible to the agent at the start of the conversation. \\
\midrule
time & The departure date, sampled uniformly from a 244-day window (2025-09-01 to 2026-05-01). \\
\midrule
context & Background context including departure city, travel dates, and group type description. Provided to both the agent and user simulators. \\
\midrule
metadata & A structured object containing: departure\_city, cities (list of 1--3 destinations), date (trip duration), group\_id, group\_name, group\_tags, compromise\_pattern (list of compromisable user IDs), and child\_members. \\
\midrule
user\_preferences & A dictionary mapping each user ID to their profile: role (e.g., ``boyfriend'', ``grandmother''), trace\_id (anonymized real profile ID), preference (hierarchical table; see \S\ref{app:preference}), and compromisable (boolean flag). \\
\midrule
initial\_messages & A list of pre-generated opening statements, one per user. Each entry contains role (user ID) and content (the statement text). Broadcast as Phase~1 of the interaction. \\
\midrule
difficulty\_score & A real-valued score: $0.5 \times f_N + 0.3 \times f_D + 0.2 \times f_C$, each dimension mapped to a 1--5 ordinal scale. \\
\midrule
difficulty\_type & One of \texttt{easy} ($s \leq 2.8$), \texttt{medium} ($2.8 < s < 4.2$), or \texttt{hard} ($s \geq 4.2$). \\
\bottomrule
\end{tabularx}
\end{table*}

\subsection{Preference Generation Strategies}
\label{app:pref_gen}

\paragraph{Design rationale: who can disagree on what.}
We generate within-group preferences under a single principle: \textbf{global constraints---budget, transport mode, activity intensity, and hotel class---are never regenerated within a related group}. This reflects how real groups actually disagree. Close-knit groups---couples, families, regular friend circles---almost never clash on whether to fly or take the train, or whether to book an economy or a comfort hotel; if they did, they typically would not be traveling together in the first place. They do, however, routinely disagree on which museum to visit or which restaurant to try. To realize this principle, we use three strategies---\emph{Independent}, \emph{Copy\_minor}, and \emph{Copy\_moderate}---assigned by a per-archetype generation plan: tight-knit archetypes (e.g., couples, parent--child families) admit only city-level conflicts, while looser archetypes (e.g., college roommates, three couples) contain more \emph{Independent} users and can additionally exhibit top-level disagreements. The resulting conflict distribution mirrors the kinds of disagreements observed in real group travel.

The three strategies (plus a \emph{Skip} option for members without independent preferences) are detailed below:

\begin{enumerate}[leftmargin=*,itemsep=2pt]
\item \textbf{Independent}: A fresh LLM call generates the complete preference table, conditioned on the user's sampled real profile. Used for the first user in each group, producing a fully original preference set.

\item \textbf{Copy\_minor}: The global constraints and city-level category preferences are inherited from a reference user (typically the partner or close companion). Only the fine-grained name-level preferences (must\_visit, reject\_visit, must\_eat, reject\_eat) are regenerated by the LLM. This simulates couples or close friends who share broad preferences but differ on specific POIs and restaurants.

\item \textbf{Copy\_moderate}: Only the top-level global constraints (budget, transport mode, intensity, hotel class) are inherited. All city-specific preferences and meso-level category preferences are regenerated. This simulates friends who share a similar travel style but have distinct interests.

\item \textbf{Skip}: For members who do not carry independent preferences (e.g., toddlers aged 0--8), no preference table is generated. These members still appear in the group and affect the \texttt{participants} field of the plan but do not contribute to utility scoring.
\end{enumerate}

After generation, each preference table undergoes three validation passes:
\begin{itemize}[leftmargin=*,itemsep=2pt]
\item \textbf{POI name validation}: Every must\_visit and must\_eat item is checked against the city's real POI inventory ($\sim$338K entries). Invalid names trigger regeneration (up to 3 retries).
\item \textbf{Cross-tier contradiction removal}: Items appearing in mutually exclusive tiers (e.g., the same POI in both must\_visit and reject\_visit) are deduplicated by keeping the stronger tier.
\item \textbf{Transport reachability check}: If any user's must/prefer transport mode has no available service on the sampled route, the user is force-flagged as compromisable ($\kappa_i = 1$).
\end{itemize}

\subsection{User Simulator}
\label{app:user_sim}

Recent benchmarks are shifting from static question answering to dynamic multi-turn interaction: fully specified tabular queries~\cite{wang2025retqa,wang2026odtqafore} give way to clarification against a user simulator~\cite{wang2026odutqa}, and we likewise withhold each user's preferences so that they surface only through multi-turn dialogue.

Each user simulator is an LLM instance with a role-conditioned system prompt that embeds:

\begin{enumerate}[leftmargin=*,itemsep=2pt]
\item The user's \textbf{role} within the group (e.g., ``You are Boyfriend A'').
\item The user's \textbf{complete preference table} in structured format.
\item A \textbf{compromise block} that varies based on the \texttt{compromisable} flag: compromisable users are instructed to ``tend to agree when the agent asks for a compromise, and append a machine-readable marker''; non-compromisable users are instructed to ``politely but firmly decline compromise requests on strong-tier preferences.''
\item \textbf{Five behavioral scenarios} defining when the user should speak: (A) when @-mentioned by the agent, (B) when a conflict with their strong preferences is detected, (C) to ask the agent factual questions, (D) to respond to a compromise request, (E) to acknowledge the agent's answer to a previous question. In all other situations, the user emits \texttt{[pass]}.
\item \textbf{Strict prohibitions}: no spontaneous preference disclosure, no planning suggestions, no @-mentioning other users, no pretending to have tool access, no repetition of previously stated preferences, no idle chat, and no answering on behalf of others.
\item \textbf{Tone mapping rules}: must = firm and non-negotiable, reject = firm refusal, prefer = mild suggestion, avoid = mild discomfort. Detailed guidelines distinguish avoid (``I'd rather not, but I can live with it'') from reject (``absolutely not'').
\end{enumerate}

The user simulator is deployed at temperature 0 with a strong instruction-following LLM to minimize behavioral variance. It sees only \emph{outward} messages (other participants' visible messages); the agent's internal reasoning, tool calls, and tool responses are hidden, just as a real user in a group chat would only see the messages, not the agent's thought process.

\subsection{Interaction Protocol Details}
\label{app:interaction_details}

\paragraph{Max turn and convergence settings by difficulty.}

\begin{table}[h]
\centering
\small
\begin{tabular}{lccc}
\toprule
& \textbf{Easy} & \textbf{Medium} & \textbf{Hard} \\
\midrule
Max turns & 15 & 20 & 25 \\
Convergence interval ($\kappa$) & 3 & 4 & 5 \\
\bottomrule
\end{tabular}
\end{table}

\paragraph{Compromise protocol mechanics.}

\begin{enumerate}[leftmargin=*,itemsep=2pt]
\item The agent identifies a conflict and @-mentions the relevant user (e.g., \texttt{@User2 Would you accept switching to high-speed rail?}).
\item The scheduler grants the @-mentioned user immediate speaking priority.
\item If the user is compromisable ($\kappa_i > 0$), the simulator replies with natural-language agreement plus a trailing machine-readable marker: \texttt{[transport.must : ["high-speed rail"]]}.
\item The framework validates that the preceding message was indeed the agent @-mentioning this user.
\item On validation success: the marker is parsed, the modification is applied to $\mathbf{p}_i^{\text{eff}}$ via dotted-path traversal, the marker is stripped from the visible message, $\kappa_i$ is decremented, and if $\kappa_i$ reaches 0, the user's system prompt is rebuilt to mark them as non-compromisable.
\item On validation failure (e.g., the marker refers to a nonexistent field path, or the preceding message was not from the agent): the user's response is regenerated (up to 3 retries).
\end{enumerate}

\paragraph{Maximum compromise quota.}
Each user can agree to at most $K = 2$ compromises. After 2 accepted compromises, the user becomes non-compromisable for the remainder of the conversation, regardless of the original \texttt{compromisable} flag. This cap prevents degenerate strategies where the agent resolves all conflicts by asking the same amenable user to yield on everything.

\paragraph{Termination conditions.}
The conversation ends when any of the following conditions is met:
\begin{enumerate}[leftmargin=*,itemsep=2pt]
\item The agent emits a structurally valid travel plan JSON (immediate termination, no revision loop).
\item The maximum number of turns is reached (the framework force-injects a final-plan instruction and the agent must generate a plan within 3 attempts).
\item A safety guard is triggered: (a) a user persistently emits \texttt{[pass]} after being @-mentioned (mention exhaustion), (b) the same (tool\_name, arguments) tuple is called $\geq 3$ times across iterations or appears in two consecutive agent turns (repetitive tool-call termination), or (c) the per-round event cap ($5 \times |\text{polling\_order}|$) is breached (runaway prevention).
\end{enumerate}

\paragraph{Subgrouping mechanics.}
When the agent decides to subgroup, the group is partitioned into $K_e$ subgroups with disjoint membership at split event $e$. Subgrouping is restricted to intra-city activities---attractions, dining, and local transportation---while shared resources remain unsplit: hotels (by night) and inter-city transportation are required to be common across all members. This restriction reflects practical group travel: members who diverge during the day still share a hotel and travel together to the next city. Each split event incurs a penalty of $K_e - 1$ in the GU computation (§\ref{sec:eval}).

\subsection{LLM-Judge Evaluation Dimensions}
\label{app:llm_judge_dims}

Table~\ref{tab:judge_dims} defines the five process-quality dimensions scored by the LLM judge. Each dimension is rated on a 1--5 ordinal scale with explicit rubric anchors. The judge is instructed to quote specific dialogue turns as evidence for each rating.

\begin{table*}[t]
\centering
\small
\setlength{\tabcolsep}{6pt}
\renewcommand{\arraystretch}{1.15}
\caption{LLM-Judge evaluation dimensions. Each dimension targets a distinct aspect of process quality that is not covered by rule-based metrics. The judge is explicitly forbidden from scoring format, time arithmetic, or numeric preference matching (handled by rules).}
\label{tab:judge_dims}
\begin{tabular}{@{}lp{12.5cm}@{}}
\toprule
\textbf{Dimension} & \textbf{Evaluation Criteria} \\
\midrule
Hallucination & Is every price, place name, and time traceable to a tool return? Does the agent fabricate information or present uncertain facts as definitive? Does it appropriately express uncertainty? \\
\midrule
Tool-Use Reasoning & Is the tool-call pattern staged and purposeful? Are calls redundant or missing? Are tool results correctly utilized in subsequent reasoning and responses? \\
\midrule
Interaction Quality & Are @-mentions used correctly and precisely? Are questions targeted, natural, and free of field-name leakage? Is dialogue pacing appropriate (not too aggressive, not too passive)? \\
\midrule
Conflict Coordination & Does the agent proactively identify preference conflicts? Is the mediation strategy explicit and equitable? Are quieter or weaker-voiced users given fair consideration? \\
\midrule
Plan Humanization & Does the daily rhythm respect meals and rest? Is there diversity across days? Are special group needs accommodated (elderly mobility, child-friendly activities)? Is the route efficient with local flavor? \\
\bottomrule
\end{tabular}
\end{table*}

\subsection{Plan Validity Check List}
\label{app:pv_checks}

The PV metric is computed by passing each generated plan through nine deterministic checks. A plan is counted as valid (PV $= 1$) only if it passes all of them; otherwise PV $= 0$.

\begin{enumerate}[leftmargin=*,itemsep=2pt]
\item \textbf{Inter-city transportation}: every required inter-city leg is covered by a flight or train segment with a feasible schedule.
\item \textbf{Hotel coverage}: every overnight stay in a destination city is matched by a hotel booking; no night is left uncovered or double-booked.
\item \textbf{Temporal consistency}: arrival times precede departure times on every segment, and no two activities overlap for the same user or subgroup.
\item \textbf{Activity overlap}: no scheduled activity (attraction visit, meal, lodging) overlaps in time with another for the same participant.
\item \textbf{Opening-hour compliance}: every attraction or restaurant is visited within its declared opening hours.
\item \textbf{Local transportation continuity}: between consecutive intra-city activities, a feasible local-transport segment exists (walking, taxi, metro, etc.).
\item \textbf{Day-level temporal monotonicity}: events within each day are in non-decreasing temporal order.
\item \textbf{Cost completeness}: each scheduled item carries a cost field consistent with the corresponding tool return.
\item \textbf{Participant validity}: every \texttt{participants} field references group members declared at task start; preschool members are not assigned tasks beyond physically reasonable activities.
\end{enumerate}

\subsection{Dataset Statistics}
\label{app:statistics}

The remaining subsections summarize the geographic, group-size, preference, and temporal distributions of the 650 released tasks.

\subsubsection{Geographic Distribution}
\label{app:geo}

The benchmark covers 143 destination cities and 32 departure cities across China. Cities are organized into a three-tier weighting system for sampling:

\begin{itemize}[leftmargin=*,itemsep=2pt]
\item \textbf{Tier-1 cities} ($\times 3$ weight, 19 cities): Beijing, Shanghai, Guangzhou, Shenzhen, Hangzhou, Chengdu, Chongqing, Wuhan, Suzhou, Xi'an, Nanjing, Changsha, Tianjin, Zhengzhou, Dongguan, Qingdao, Kunming, Ningbo, Foshan.
\item \textbf{Popular tourist cities} ($\times 2$ weight, $\sim$55 cities): e.g., Xiamen, Guilin, Lijiang, Sanya, Huangshan, Dali, Luoyang.
\item \textbf{General tourist cities} ($\times 1$ weight, $\sim$80 cities): remaining cities with notable tourist resources.
\end{itemize}

Destination cities are further organized into 175 geographically coherent multi-city clusters (derived from 10 regional tourism clusters: North China, Yangtze River Delta, Pearl River Delta, Southwest, etc.), ensuring that multi-city trips follow realistic travel routes rather than arbitrary city combinations.

\begin{figure}[t]
\centering
\includegraphics[width=\columnwidth]{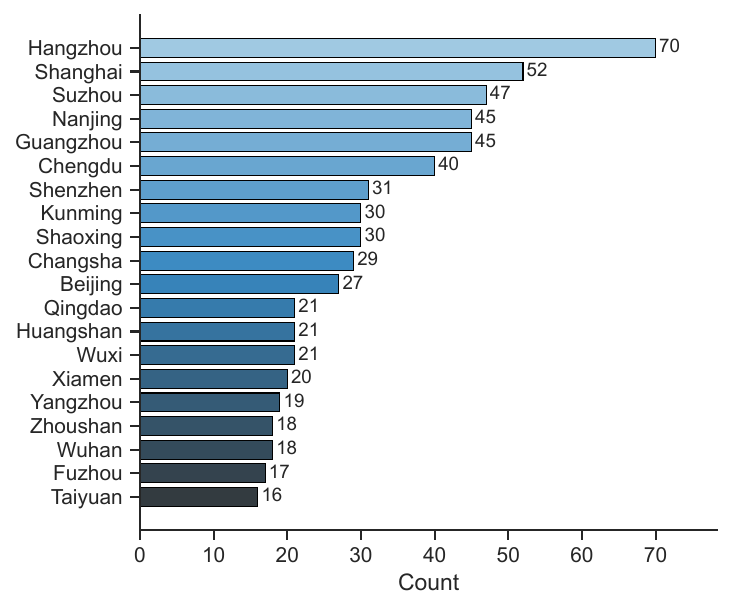}
\caption{Top-20 destination city frequency in the 650-task benchmark.}
\label{fig:city_dist}
\end{figure}

The 32 departure cities correspond to provincial capitals and major transportation hubs, ensuring realistic intercity connectivity. The most frequent departure cities include Guiyang (31), Hefei (31), Shanghai (28), Fuzhou (26), and Changsha (26). Figure~\ref{fig:city_dist} shows the top-20 most frequent destination cities, reflecting the tier-weighted sampling strategy: Tier-1 cities dominate due to their $\times 3$ weight, while popular tourist cities form the long tail.

\subsubsection{Group Size and Difficulty Distribution}
\label{app:difficulty}

Table~\ref{tab:difficulty_cross} shows the cross-tabulation of group size, city count, and trip duration against difficulty level. The difficulty score is computed as a weighted combination:
\begin{equation}
s = 0.5 \times f_N(N) + 0.3 \times f_D(D) + 0.2 \times f_C(C),
\end{equation}
where $f_N$, $f_D$, $f_C$ map group size $N$, duration $D$ (in days), and city count $C$ to $\{1,\ldots,5\}$ ordinal scales respectively.

\begin{table}[t]
\centering
\caption{Cross-tabulation of key task dimensions by difficulty level.}
\label{tab:difficulty_cross}
\small
\begin{tabular}{lccc}
\toprule
& \textbf{Easy (200)} & \textbf{Medium (250)} & \textbf{Hard (200)} \\
\midrule
\multicolumn{4}{l}{\textit{Group size ($N$)}} \\
\quad 2 & 65 & 9 & 0 \\
\quad 3 & 77 & 44 & 0 \\
\quad 4 & 42 & 84 & 0 \\
\quad 5 & 16 & 33 & 75 \\
\quad 6 & 0 & 80 & 125 \\
\midrule
\multicolumn{4}{l}{\textit{City count}} \\
\quad 1 & 116 & 43 & 0 \\
\quad 2 & 82 & 131 & 48 \\
\quad 3 & 2 & 76 & 152 \\
\midrule
\multicolumn{4}{l}{\textit{Duration}} \\
\quad 2--3 days & 144 & 66 & 0 \\
\quad 4--5 days & 56 & 134 & 109 \\
\quad 6--7 days & 0 & 50 & 91 \\
\bottomrule
\end{tabular}
\end{table}

As shown, easy tasks are dominated by 2--3 person groups visiting a single city for 2--3 days, while hard tasks exclusively feature 5--6 person groups across 2--3 cities over 5--7 days. This design ensures that difficulty scales primarily along the \emph{coordination axis}: larger groups produce more pairwise preference conflicts, more cities increase planning complexity, and longer durations expand the constraint space.

\subsubsection{Preference Distribution}
\label{app:pref_dist}

Table~\ref{tab:pref_dist} summarizes the distribution of preference items across the 2,748 users in the benchmark.

\begin{table*}[t]
\centering
\caption{Preference field statistics across all 2,748 users.}
\label{tab:pref_dist}
\small
\setlength{\tabcolsep}{10pt}
\renewcommand{\arraystretch}{1.15}
\begin{tabular}{lr}
\toprule
\textbf{Statistic} & \textbf{Value} \\
\midrule
\multicolumn{2}{@{}l}{\textit{Global constraints}} \\
\quad Budget constraint coverage & 100\% \\
\quad Intensity constraint coverage & 100\% \\
\quad Transport must / prefer / avoid / reject & 2,184 / 2,230 / 2,172 / 182 \\
\quad Hotel prefer / avoid & 5,469 / 2,741 \\
\midrule
\multicolumn{2}{@{}l}{\textit{City-specific preferences}} \\
\quad Attractions: must\_visit / reject\_visit & 7,699 / 2,501 \\
\quad Attractions: category pos. / neg. & 7,195 / 4,460 \\
\quad Food: must / prefer / avoid / reject & 5,945 / 5,940 / 3,896 / 2,140 \\
\midrule
\multicolumn{2}{@{}l}{\textit{Aggregates}} \\
\quad Total atomic preference points & 62,998 \\
\quad Average per user & 22.9 \\
\quad Compromisable users & 54.2\% (1,490 / 2,748) \\
\quad Budget range (per person) & 550--10,500 CNY \\
\quad Budget mean / median & 3,265 / 3,000 CNY \\
\bottomrule
\end{tabular}
\end{table*}

Every user carries both a budget cap and an intensity cap, ensuring that every task has quantitative constraints the agent must discover and satisfy. Transport preferences are roughly balanced between positive (must: 2,184, prefer: 2,230) and avoidance (avoid: 2,172) tiers, with hard rejections (reject: 182) being deliberately rare to avoid trivially infeasible tasks. City-specific preferences are richer: an average user has $\sim$2.8 must-visit attractions, $\sim$0.9 reject-visit attractions, $\sim$2.6 positive categories, and $\sim$1.6 negative categories per city, plus $\sim$2.2 must-eat and $\sim$2.2 prefer-eat food items.

\paragraph{Intra-group preference conflicts.}
We measure conflict density at the \emph{all-tier} level: a conflict is recorded whenever the same item appears in one user's positive tier (\texttt{must} / \texttt{prefer} / \texttt{category\_pref.positive}) and another user's negative tier (\texttt{reject} / \texttt{avoid} / \texttt{category\_pref.negative}), within the same task and the same city when applicable. Conflicts are counted at the item level across all directed user pairs. Under this definition, 503 / 650 tasks (77.4\%) contain at least one preference conflict, with an overall mean of 7.52 conflicts per task (median 4, max 54). Broken down by conflict source:

\begin{itemize}[leftmargin=*,itemsep=2pt]
\item \textbf{Food}: 365 / 650 tasks (56.2\%) -- the most common source, reflecting the breadth of must/prefer/avoid/reject options in the food schema.
\item \textbf{Attraction category}: 282 / 650 tasks (43.4\%) -- positive vs.\ negative category-level disagreements.
\item \textbf{Attraction name}: 215 / 650 tasks (33.1\%) -- direct \texttt{must\_visit} vs.\ \texttt{reject\_visit} clashes on specific POIs.
\item \textbf{Transport}: 165 / 650 tasks (25.4\%) -- positive vs.\ negative tier disagreements on transport mode.
\item \textbf{Hotel}: 151 / 650 tasks (23.2\%) -- \texttt{prefer} vs.\ \texttt{avoid} clashes on hotel categories.
\end{itemize}

\begin{figure}[t]
\centering
\includegraphics[width=\columnwidth]{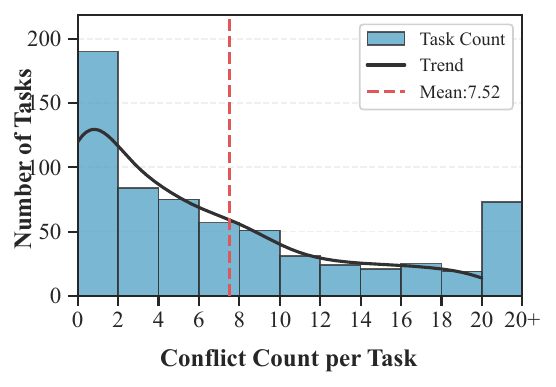}
\caption{Distribution of per-task all-tier preference conflict counts across the 650 tasks. Counts above 20 are aggregated into the rightmost ``20+'' bucket.}
\label{fig:conflict_dist}
\end{figure}

Conflict density scales sharply with difficulty: easy tasks average 1.82 conflicts (48.5\% with any conflict), medium tasks average 6.52 (84.0\%), and hard tasks average 14.46 (98.0\%). Figure~\ref{fig:conflict_dist} shows the full per-task distribution: most tasks fall below 10 conflicts but a heavy right tail extends past 20, ensuring substantial coverage for evaluating conflict-discovery and coordination abilities. Hard-tier clashes (\texttt{must} vs.\ \texttt{reject}) remain deliberately rare on transport (5 / 650, 0.8\%) so as not to render tasks infeasible.

\subsubsection{Temporal Distribution}
\label{app:temporal}

Departure dates span a 244-day window from September 1, 2025 to May 1, 2026, covering:
\begin{itemize}[leftmargin=*,itemsep=2pt]
\item National Day holiday (Oct 1--7, 2025)
\item Winter holiday and Spring Festival season (Jan--Feb 2026)
\item Qingming, Labor Day (Apr--May 2026)
\item Regular weekdays and weekends throughout the period
\end{itemize}

This temporal range ensures that opening-hours checks, weather queries, and seasonal pricing in the sandbox reflect realistic variation, providing additional test surface beyond static preference matching.

\section{Sandbox Tools and Cache}
\label{app:tools}

To support stable, reproducible offline evaluation, the agent interacts with a sandboxed tool environment rather than live APIs. This appendix documents the tool library, the cache that backs it, the fallback mechanism used on cache misses, the call schema seen by the model, and the empirical distribution of cached entries across tools.

\subsection{Tool Library}
\label{app:tool_library}

The agent has access to 10 real, production-grade travel tools organized into four domains. Table~\ref{tab:tools} provides an overview, and detailed descriptions follow.

\begin{table*}[t]
\centering
\small
\setlength{\tabcolsep}{6pt}
\renewcommand{\arraystretch}{1.15}
\caption{Overview of the tool library used in our benchmark sandbox, grouped by domain.}
\label{tab:tools}
\begin{tabular}{lll}
\toprule
\textbf{Domain} & \textbf{Tool name} & \textbf{Function} \\
\midrule
\multirow{6}{*}{Maps \& navigation}
& \texttt{search\_poi} & Retrieve POIs by keyword, coordinates, or area \\
& \texttt{get\_poi\_detail} & Retrieve detailed POI information by ID (hours, reviews, pricing) \\
& \texttt{maps\_geo} & Geocode an address or place name to coordinates \\
& \texttt{plan\_route} & Plan routes across modes from origin to destination \\
& \texttt{compare\_routes} & Compare route options across different transport modes \\
& \texttt{search\_along\_route} & Search POIs within a corridor along a given route \\
\midrule
\multirow{2}{*}{Transportation}
& \texttt{travel\_search\_flights} & Search China domestic flights with flexible dates \\
& \texttt{travel\_search\_trains} & Search China train and high-speed rail trips with flexible dates \\
\midrule
Weather
& \texttt{maps\_weather} & Return current weather and multi-day forecasts \\
\midrule
General information
& \texttt{web\_search} & Perform open-domain web search \\
\bottomrule
\end{tabular}
\end{table*}

\paragraph{Maps \& Navigation Tools.}

\begin{enumerate}[leftmargin=*,itemsep=3pt]
\item \texttt{search\_poi}: A large-coverage POI retrieval tool supporting nationwide search across China. It accepts keywords, categories, city names, and coordinate-based radius queries. Results include structured metadata: name, address, latitude/longitude, category taxonomy (AMap's standardized classification), user rating, price level, opening hours, and review count. The tool supports multiple ranking strategies (distance, rating, composite).

\item \texttt{get\_poi\_detail}: Given a POI ID returned by \texttt{search\_poi}, retrieves comprehensive details including full opening hours (structured with seasonal and weekday-specific rules), user review excerpts, high-resolution photos, contact information, and pricing signals. This tool is essential for the Plan Validity check on opening-hours compliance.

\item \texttt{maps\_geo}: Geocodes a free-form address or place name into latitude/longitude coordinates. Used as a preprocessing step for distance-based queries and route planning.

\item \texttt{plan\_route}: Computes routes between an origin and destination, each specifiable as a free-form address or explicit coordinates. Supports four transportation modes: driving, walking, cycling, and public transit. Returns route summaries (total distance, estimated duration, traffic conditions) and step-by-step navigation instructions. Supports practical constraints such as toll avoidance and highway preference.

\item \texttt{compare\_routes}: A convenience wrapper that computes routes across multiple transport modes between the same origin--destination pair, enabling side-by-side comparison of duration, distance, and cost.

\item \texttt{search\_along\_route}: Searches for POIs within a user-specified buffer corridor along a planned route. Useful for requests such as ``find a coffee shop near my route'' or ``find a rest stop along the highway.'' The tool first plans a base route, then searches for POIs within the buffer region.
\end{enumerate}

\paragraph{Transportation Tools.}

\begin{enumerate}[leftmargin=*,itemsep=3pt]
\item \texttt{travel\_search\_flights}: Searches domestic flight options between two Chinese cities. Supports flexible date ranges (multi-day queries). Returns structured results: flight number, airline, departure/arrival times, aircraft type, and price ranges by cabin class.

\item \texttt{travel\_search\_trains}: Queries conventional train and high-speed rail services between two cities. Supports multi-day flexible date queries. Returns train number, departure/arrival stations, intermediate stops, travel duration, and ticket prices by seat class.
\end{enumerate}

\paragraph{Weather and General Information.}

\begin{enumerate}[leftmargin=*,itemsep=3pt]
\item \texttt{maps\_weather}: Retrieves current weather conditions (temperature, feels-like, wind, precipitation, phenomena) and multi-day forecasts (up to 5 days) for a specified location. Supports both single-date and date-range queries.

\item \texttt{web\_search}: Performs open-domain web search for information that falls outside the scope of the domain-specific tools, such as local regulations, travel policies, seasonal events, or cultural information.
\end{enumerate}

\subsection{Cache System}
\label{app:cache}

To ensure \textbf{stable and reproducible evaluation}, all tool calls are routed through a content-addressable cache system. The cache operates in two modes:

\paragraph{ISOLATED mode (default for evaluation).}
During evaluation, the sandbox operates in \texttt{ISOLATED} mode. All tool invocations are resolved from a pre-built cache:

\begin{enumerate}[leftmargin=*,itemsep=2pt]
\item The tool call's normalized argument JSON is used as the cache key.
\item If an exact match exists, the cached response is returned immediately.
\item If no exact match exists (cache miss), the system falls back to the \textbf{embedding-retrieval + ICL simulation} strategy (\S\ref{app:cache_miss}).
\end{enumerate}

This ensures that identical agent trajectories on identical cached states produce identical tool outputs, eliminating environmental variance from the evaluation.

\paragraph{ONLINE mode (for cache warming).}
During the initial cache-building phase, the sandbox operates in \texttt{ONLINE} mode. Tool calls first check the cache; on miss, they are forwarded to the real API endpoint (AMap, flight/train APIs, web search). The response is cached for future use. This mode is used only during dataset construction and cache warming, never during evaluation.

\paragraph{Cache implementation details.}
\begin{itemize}[leftmargin=*,itemsep=2pt]
\item Each tool has a dedicated JSON cache file, avoiding cross-tool key collisions.
\item Cache writes are thread-safe and use atomic file operations (write to temp file, then rename) to prevent corruption under concurrent access.
\item An auto-save mechanism flushes dirty entries to disk every 30 seconds.
\item Cache misses during evaluation are logged to separate \texttt{*\_missed.json} files for later batch refresh.
\end{itemize}

\subsection{Cache-Miss Simulation}
\label{app:cache_miss}

Despite pre-warming the cache with extensive crawling, exact-match misses are unavoidable due to minor variations in tool arguments (e.g., ``Beijing'' vs.\ ``Beijing'', different radius values, slightly different coordinate precision). To handle these misses deterministically, we employ an \textbf{embedding-based retrieval + in-context learning (ICL)} strategy, following the approach validated in TravelBench~\cite{cheng2025travelbench}:

\begin{enumerate}[leftmargin=*,itemsep=2pt]
\item \textbf{Embedding precomputation}: We precompute embeddings for all cached tool-call inputs using Qwen3-Embedding-8B, deployed as a remote embedding service. Embeddings are stored as \texttt{.npz} files alongside each tool's cache.

\item \textbf{FAISS-based retrieval}: When a cache miss occurs, the current tool call's input is embedded and used to query a FAISS index (built at startup) to retrieve the top-8 most similar cached entries.

\item \textbf{ICL simulation}: The retrieved (input, output) pairs are formatted as few-shot examples, along with the tool's schema definition. A tool-simulator LLM generates a plausible response that is consistent with the real tool's output format and the retrieved examples.

\item \textbf{Transparent logging}: Simulated responses are saved to the missed-calls file but are \emph{not} written back to the main cache, ensuring that the primary cache remains a faithful record of real API responses.
\end{enumerate}

This strategy ensures that (1) the simulated response distribution stays close to the real tool's output distribution, (2) evaluation remains deterministic for the same embedding model and simulator LLM, and (3) researchers can later refresh the cache with real API calls by replaying the missed-calls log.

\subsection{Tool Call Schema Format}
\label{app:tool_schema}

All tools are registered with the agent in OpenAI function-calling format. Each tool definition includes:
\begin{itemize}[leftmargin=*,itemsep=2pt]
\item \texttt{name}: the tool identifier (e.g., \texttt{search\_poi})
\item \texttt{description}: a natural-language description of the tool's purpose and capabilities
\item \texttt{parameters}: a JSON Schema object defining required and optional parameters, their types, enumerations, and descriptions
\end{itemize}

The complete schema definitions for all 10 tools are included in the released codebase. During evaluation, tool invocations undergo strict schema validation: required-field checks, type constraints, and range constraints are enforced. Calls that fail validation are recorded as tool-call errors and are reflected in the execution statistics.

\subsection{Cache Distribution}
\label{app:cache_dist}

\begin{figure}[t]
\centering
\includegraphics[width=\columnwidth]{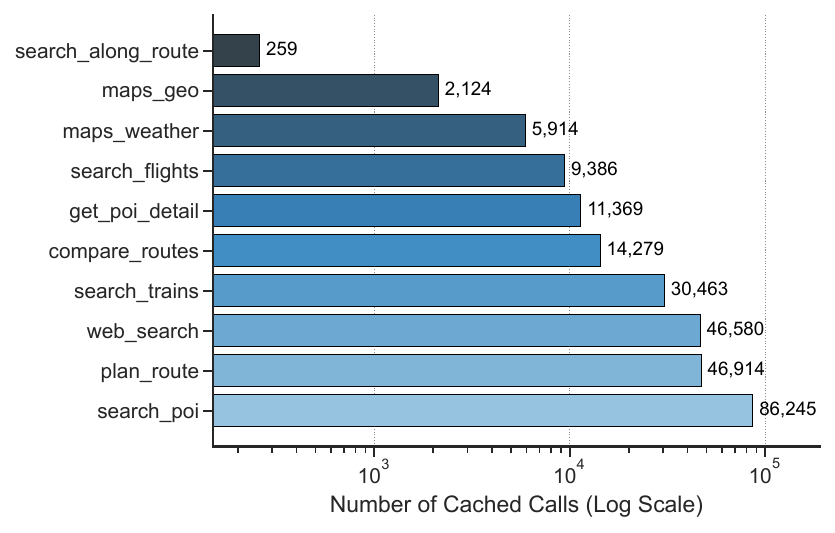}
\caption{Distribution of cached entries across the 10 sandbox tools. Total: 253,533 entries.}
\label{fig:cache_dist}
\end{figure}

Figure~\ref{fig:cache_dist} shows the distribution of pre-warmed cache entries across the 10 tools. \texttt{search\_poi} dominates with 86K entries, reflecting the core role of POI retrieval in travel planning. Route-related tools (\texttt{plan\_route}, \texttt{web\_search}) each have $\sim$47K entries, while niche tools (\texttt{search\_along\_route}: 259 entries) require the embedding-retrieval fallback more frequently.

\section{Additional Experiments}
\label{app:exp_details}

This appendix supplements the main experiments with implementation details, model exclusion criteria, and finer-grained behavioral analyses (group-size scaling, compromise/tool-use patterns, plan-validity error breakdown, judge calibration, and per-model distributions) that the main text could not fit.

\subsection{Implementation Details}
\label{app:implementation}

\paragraph{Agent and simulator deployment.}
For each evaluated agent, we set the sampling temperature to 0.7 and run $T=3$ independent trials per task, reporting the mean score across trials. The maximum output length for both agent and user simulator responses is set to 8{,}192 tokens. The user simulator uses DeepSeek-V4-Flash at temperature 0 to minimize behavioral variance.

\paragraph{Open-source model serving.}
All open-source models (Qwen3-235B-A22B-Thinking, Qwen3.5-Plus, Qwen3.6-Max-Preview, Qwen3.5-Flash, Qwen3-30B-A3B-Thinking, Qwen3-4B-Thinking, and Qwen3-4B-Instruct) are deployed via vLLM with the maximum sequence length set to 131{,}072 tokens to accommodate long multi-turn conversations with tool-call traces. We enable tensor parallelism as needed for larger models. For thinking models, we enable the extended thinking mode with \texttt{enable\_thinking=true}.

\paragraph{Embedding model.}
For cache-miss retrieval in the tool sandbox (\S\ref{app:cache_miss}), we deploy Qwen3-Embedding-8B with vLLM as a remote embedding service. We precompute embeddings for all cached tool-call inputs and store them as compressed \texttt{.npz} files. At evaluation time, FAISS \texttt{IndexFlatIP} indices (cosine similarity with $L_2$-normalized vectors) are built from these precomputed embeddings to enable fast top-8 similarity retrieval.

\paragraph{Tool simulator.}
When a cache miss occurs in \texttt{ISOLATED} mode, the tool simulator LLM generates a plausible response using retrieved examples as few-shot context. We use DeepSeek-V4-Flash as the tool simulation LLM with temperature 0, ensuring stable and deterministic simulation results.

\paragraph{LLM-as-Judge.}
We use Gemini-3-Flash-Preview as the judge model for both primary evaluation and meta-judging, with temperature set to 0. Each task is evaluated independently across its $T=3$ trials, producing 1{,}950 evaluation samples per model (650 tasks $\times$ 3 trials). The meta-judge reviews the primary judge's reasoning and produces a calibration score $s^{\text{meta}} \in \{1,\ldots,5\}$, which scales the final LLM-Judge scores.

\paragraph{Difficulty-aware hyperparameters.}
Max interaction rounds and convergence intervals are adapted to task difficulty: Easy tasks allow 15 rounds with convergence interval $\kappa=3$, Medium tasks allow 20 rounds with $\kappa=4$, and Hard tasks allow 25 rounds with $\kappa=5$. These settings ensure that harder tasks (larger groups, more cities) receive proportionally more interaction budget.

\paragraph{Sandbox modes.}
All main results are evaluated in \texttt{ISOLATED} mode, where tool calls are resolved entirely from the pre-built cache (253{,}533 entries) with embedding-retrieval fallback for misses. We additionally evaluate a subset of models in \texttt{ONLINE} mode (live API calls) to validate sandbox fidelity (\S\ref{sec:offline_online}).

\subsection{Excluded Models}
\label{app:excluded_models}

In addition to the models reported in the main results, we tested several additional models that ultimately failed to complete the benchmark reliably:

\begin{itemize}[leftmargin=*,itemsep=3pt]
\item \textbf{MiniMax-M2.5}~\cite{minimax_2025}: This model produces a large number of invalid tool-call arguments, especially malformed JSON outputs. Common issues include incorrect JSON structure, missing required fields, and arguments that do not conform to the expected schema. As a result, a substantial fraction of its tool calls fail validation, preventing reliable evaluation.

\item \textbf{Kimi-K2.5}~\cite{team2025kimi_k2}: This model frequently fails to generate valid tool calls, often producing incorrect tool names or invalid parameters. It also exhibits repeated outputs, including repeated tool-call attempts with the same or similarly malformed content, which prevents the interaction from making meaningful progress.

\item \textbf{Qwen3-30B-A3B-Instruct} and \textbf{Qwen3-235B-A22B-Instruct}: Both instruction-following variants of the Qwen3 family suffer from persistent repetitive tool-call loops, repeatedly issuing the same search queries without advancing the conversation. A large fraction of tasks for both variants terminate early due to repeated tool calls.

\item \textbf{Llama3} and \textbf{Llama3.3}: These models are generally unable to produce executable tool calls under our benchmark setting. In most cases, they fail to generate the required tool-call format altogether, making them unusable for our evaluation pipeline.
\end{itemize}

These failures reveal a common challenge: many models struggle to maintain valid and effective tool use in long-horizon multi-party planning. In particular, instruction-following models without explicit reasoning often fail to execute coherent multi-step tool-use strategies over extended conversations. In contrast, their thinking-mode counterparts (Qwen3-30B-Thinking and Qwen3-235B-Thinking) achieve near-100\% plan generation rates, suggesting that explicit reasoning is important for sustained agentic behavior. Notably, GPT-5.1, despite being an instruction-following model, achieves a 96.4\% plan generation rate, indicating that strong model capability can partially compensate for the lack of explicit chain-of-thought in this setting.

\subsection{Group-Size Scaling}
\label{app:scaling}

\begin{figure*}[t]
\centering
\includegraphics[width=\textwidth]{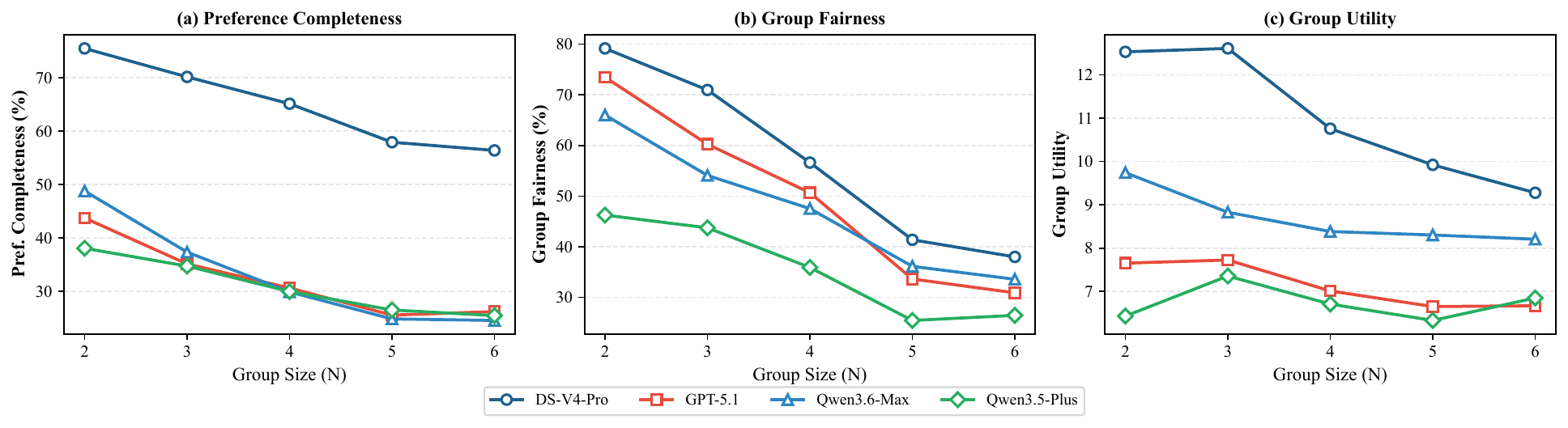}
\caption{Performance scaling with group size ($N=2\dots6$). Preference Completeness and Group Fairness degrade sharply, while Group Utility increases---multi-party coordination, not planning capacity, is the core bottleneck.}
\label{fig:scaling}
\end{figure*}

Figure~\ref{fig:scaling} isolates the effect of group size ($N=2,\dots,6$) on the three core abilities. Preference Completeness drops steeply as $N$ grows: DeepSeek-V4-Pro falls from 76\% ($N=2$) to 57\% ($N=6$), while other models converge to $\sim$25\%. Group Fairness shows an even sharper decline---from $\sim$75\% at $N=2$ to $\sim$35\% at $N=6$ across all models---confirming that coordination difficulty scales super-linearly with the number of stakeholders. In contrast, Group Utility \emph{increases} with $N$, because harder tasks involve more cities and days, offering more scheduling slots. This divergence between GU and GF reveals the core challenge: models can accumulate aggregate utility by satisfying the vocal majority, but fail to distribute it fairly when the number of voices grows.

\subsection{Compromise Behavior}
\label{app:compromise_behavior}

Table~\ref{tab:model_stats} summarizes interaction characteristics. DeepSeek-V4-Pro triggers the most compromises (0.65/task) and uses the most tools (33.8/task), reflecting an aggressive strategy that pays off in higher utility. GPT-5.1 uses more interaction rounds (8.9) but triggers fewer compromises (0.26), suggesting it prefers to gather information over resolving conflicts. Thinking models (Qwen3-30B, Qwen3-4B) use almost no tools ($<4$) and barely interact (0.5--2.3 rounds), essentially generating plans from a single pass.

\begin{table}[t]
\small
\centering
\setlength{\tabcolsep}{4pt}
\renewcommand{\arraystretch}{0.95}
\caption{Model-level interaction statistics. \textit{Comp.} = avg compromises per task, \textit{Tools} = avg tool calls, \textit{Rnds} = avg rounds used, \textit{Plan\%} = plan generation rate.}
\label{tab:model_stats}
\begin{tabular}{lcccc}
\toprule
\textbf{Model} & \textbf{Comp.} & \textbf{Tools} & \textbf{Rnds} & \textbf{Plan\%} \\
\midrule
DeepSeek-V4-Pro & 0.65 & 33.8 & 7.3 & 99.7 \\
GPT-5.1         & 0.26 & 12.3 & 8.9 & 96.4 \\
Qwen3.6-Max     & 0.30 & 20.0 & 2.2 & 87.0 \\
Qwen3.5-Plus    & 0.38 & 23.8 & 4.3 & 73.2 \\
Qwen3-235B-Th   & 0.30 & 8.6 & 3.3 & 99.9 \\
Qwen3.5-Flash   & 0.25 & 20.3 & 1.3 & 91.2 \\
Qwen3-30B-Th    & 0.31 & 3.6 & 0.5 & 100.0 \\
Qwen3-4B-Th     & 0.20 & 3.3 & 2.3 & 100.0 \\
Qwen3-4B-It     & 0.15 & 30.1 & 4.2 & 89.0 \\
\bottomrule
\end{tabular}
\vspace{-8pt}
\end{table}

\begin{figure}[t]
\centering
\includegraphics[width=\columnwidth]{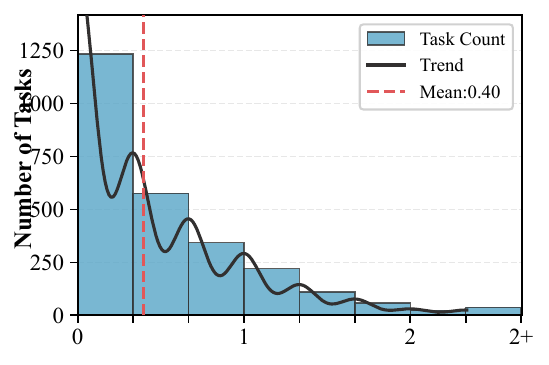}
\caption{Per-task compromise count distribution (averaged over 4 models $\times$ 3 trials). Most tasks see zero compromises; the negotiation mechanism is underutilized.}
\label{fig:compromise_dist}
\end{figure}

Figure~\ref{fig:compromise_dist} shows the per-task compromise distribution averaged across four representative models. The majority of tasks (67\%) receive zero successful compromises on average, with only 2\% reaching two or more. This reveals that the compromise mechanism---while available---is severely underutilized: most models skip conflict negotiation and proceed directly to planning, accepting lower fairness scores as a consequence.

Two compounding factors likely drive this gap. First, the agents' own conflict-identification and resolution skills are limited: even when a conflict is in plain sight, the model often plans around it rather than negotiating. Second, preference elicitation itself is incomplete---when an agent's inferred preference table $\hat{\mathbf{p}}_i$ is too sparse (Preference Completeness stays at or below 64\% for the strongest model in Table~\ref{tab:main}), the agent cannot detect a conflict in the first place, so the compromise mechanism never has a chance to fire.

\subsection{Tool-Use Patterns}
\label{app:tool_use_patterns}

\begin{figure}[t]
\centering
\includegraphics[width=\columnwidth]{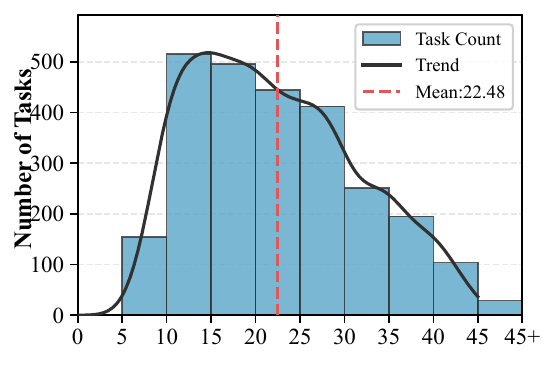}
\caption{Per-task tool call distribution (averaged over 4 models $\times$ 3 trials). The near-normal shape (mean=22.5) confirms consistent grounding demand.}
\label{fig:tool_dist}
\end{figure}

Figure~\ref{fig:tool_dist} shows the per-task tool call distribution (averaged across four models). The approximately normal distribution (mean=22.5, range 10--35) confirms that multi-person travel planning demands consistent, substantial tool grounding---models cannot produce valid plans from parametric knowledge alone. The moderate spread reflects variation in task complexity (more cities and days require more queries).

\subsection{Plan Validity Error Analysis}
\label{app:validity_analysis}

\begin{figure*}[t]
\centering
\includegraphics[width=0.85\textwidth]{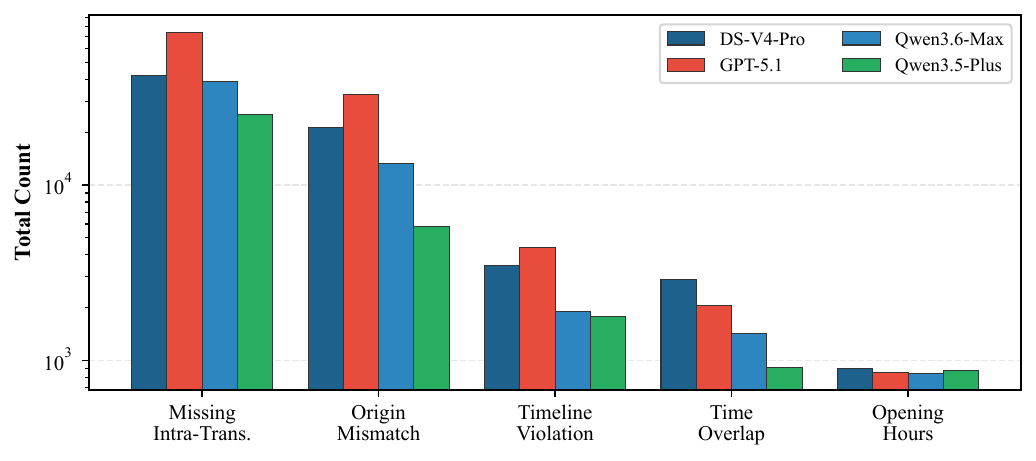}
\caption{Plan validity error breakdown by type, 4 models (log scale). Missing intra-city transport dominates by an order of magnitude, identifying spatial-continuity reasoning as the primary structural failure mode.}
\label{fig:validity}
\end{figure*}

Figure~\ref{fig:validity} breaks down structural error types across four models (log scale). Two error categories dominate orders-of-magnitude above others: \emph{missing intra-city transport} (25K--74K instances) and \emph{transport origin mismatch} (6K--33K). This reveals that maintaining spatial continuity---ensuring every location transition is covered by an explicit transport activity---is the single hardest structural challenge. Even DeepSeek-V4-Pro, the best-performing model, accumulates 42K missing-transport errors across its 1,950 samples. The remaining error types (timeline violations, time overlaps, opening-hours conflicts) are 5--50$\times$ less frequent, suggesting that temporal reasoning is comparatively easier than spatial-continuity tracking.

\subsection{LLM-Judge Human Validation}
\label{app:judge_human_validation}

While the meta-judge calibration (Appendix~\ref{app:meta_judge}) corrects individual judge verdicts at the trajectory level, it does not directly establish that the LLM-judge scores correspond to human preferences. We therefore conducted a direct human validation study.

\paragraph{Setup.}
From each of the four models in Figure~\ref{fig:radar} (DeepSeek-V4-Pro, GPT-5.1, Qwen3.6-Max, and Qwen3.5-Plus), we randomly sampled \textbf{25 trajectories}, yielding \textbf{100 trajectories} in total. Three independent annotators scored each trajectory under the same five-dimensional rubric used by the LLM judge (Appendix~\ref{app:llm_judge_dims}), assigning discrete scores from 1 to 5.

\paragraph{Metrics.}
We report mean absolute error (MAE) in two forms:
(1) \textbf{MAE(LLM vs.\ human)} is the MAE between the LLM judge's score and each annotator's score, averaged across annotators;
(2) \textbf{MAE(human vs.\ human)} uses the median of the three human annotations as a reference, computes each annotator's MAE to this median, and then averages the results. Because trajectory evaluation is inherently subjective, this human--human deviation serves as an approximate upper bound on the agreement one can expect.

\begin{table}[t]
\centering
\small
\begin{tabular}{lcc}
\toprule
 & LLM--Hum. & Hum.--Hum. \\
\midrule
Avg.\ (5 dims.) & 0.77 & 0.62 \\
\bottomrule
\end{tabular}
\caption{Agreement between the LLM judge and human annotators on 100 sampled trajectories (25 per model). \textbf{LLM--Hum.}\ is MAE between the LLM judge and individual annotators, averaged over annotators. \textbf{Hum.--Hum.}\ is MAE between each annotator and the median of the three human ratings, averaged over annotators. \textbf{Avg.\ (5 dims.)}\ is the average over the five evaluation dimensions.}
\label{tab:judge_human_agreement}
\end{table}

\paragraph{Results.}
As shown in Table~\ref{tab:judge_human_agreement}, scoring multi-turn group-planning trajectories is inherently subjective: even between human annotators, the average MAE is \textbf{0.62} on the 1--5 scale---well above zero, reflecting the open-ended nature of judging behaviors such as preference elicitation, conflict mediation, and itinerary humanization. Against this backdrop, the LLM judge incurs an average MAE of \textbf{0.77} versus individual annotators, only \textbf{0.15} higher than the human--human ceiling. We therefore conclude that the LLM judge introduces only a small amount of additional disagreement beyond the task's inherent subjectivity, and its scores remain within a controllable range relative to human raters.

\subsection{Meta-Judge Calibration}
\label{app:meta_judge}

Table~\ref{tab:meta_judge} summarizes the meta-judge score distribution across all 15,596 evaluated samples (8 models $\times$ $\sim$1,950 samples each). 98.4\% of evaluations receive the maximum calibration score (5/5), indicating that the primary judge's reasoning is almost always well-grounded and does not require adjustment. Only 0.4\% of samples score $\leq$3, confirming that the meta-judge serves as a lightweight quality filter rather than a frequent correction mechanism.

\begin{table}[t]
\small
\centering
\setlength{\tabcolsep}{8pt}
\renewcommand{\arraystretch}{0.95}
\caption{Meta-judge score distribution (15,596 total samples).}
\label{tab:meta_judge}
\begin{tabular}{cccccc}
\toprule
\textbf{Score} & 1 & 2 & 3 & 4 & 5 \\
\midrule
\textbf{Count} & 2 & 15 & 41 & 187 & 15,351 \\
\textbf{Ratio} & 0.0\% & 0.1\% & 0.3\% & 1.2\% & 98.4\% \\
\bottomrule
\end{tabular}
\vspace{-8pt}
\end{table}

\subsection{Completion Reason Breakdown}
\label{app:completion_reasons}

\begin{figure*}[t]
\centering
\includegraphics[width=0.75\textwidth]{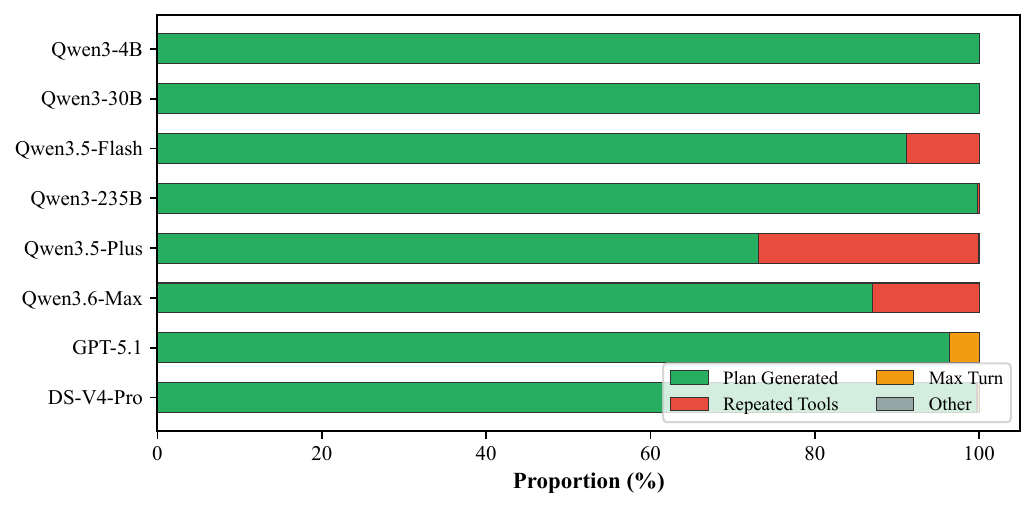}
\caption{Completion reason distribution. Repetitive tool-call termination (red) disproportionately affects Qwen3.5-Plus and Qwen3.6-Max, while thinking models achieve near-100\% plan generation.}
\label{fig:completion_reasons}
\end{figure*}

Figure~\ref{fig:completion_reasons} shows the completion reason distribution across all eight models. Models fall into three categories: (1)~\emph{reliable finishers} (DS-V4-Pro, Qwen3-235B, Qwen3-30B, Qwen3-4B) that almost always emit a valid plan; (2)~\emph{mostly successful} (GPT-5.1, Qwen3.5-Flash) with 4--9\% failures due to max-turn or repeated tool calls; (3)~\emph{frequently failing} (Qwen3.5-Plus, Qwen3.6-Max) where 13--27\% of runs terminate due to repetitive tool-call loops. The repetitive-tool-call termination mode is unique to instruction-following models that lack explicit reasoning chains.

\subsection{LLM-Judge Sub-Dimension Comparison (All Models)}
\label{app:subdim_bar}

\begin{figure*}[t]
\centering
\includegraphics[width=0.85\textwidth]{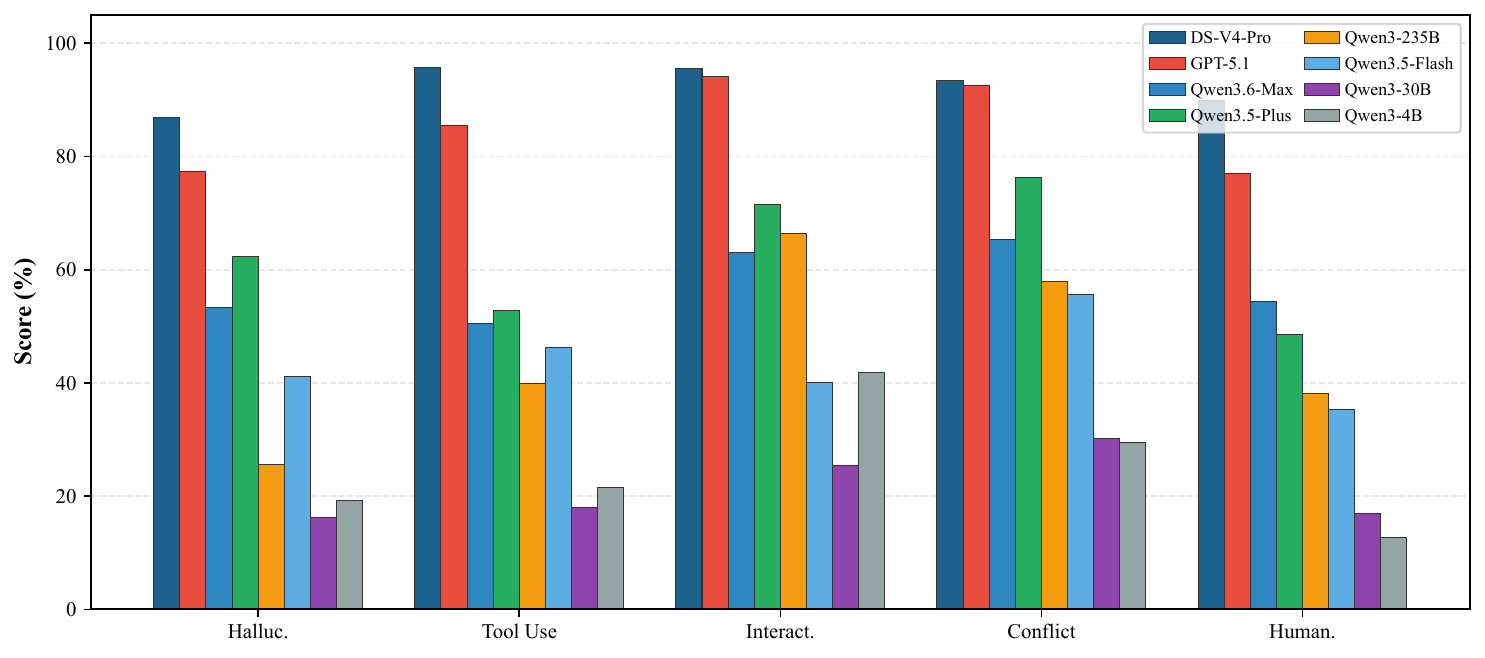}
\caption{LLM-Judge sub-dimension scores for all 8 models. Hallucination shows the widest inter-model spread; Interaction Quality is the most uniformly distributed.}
\label{fig:subdim_bar}
\end{figure*}

Figure~\ref{fig:subdim_bar} provides a grouped bar chart comparing all eight models across the five LLM-Judge dimensions. The ``Hallucination'' dimension shows the widest spread (16--87\%), confirming it as the primary differentiator between models. ``Interaction Quality'' has the narrowest gap between top and bottom models, suggesting that basic dialogue competence is more uniformly distributed.

\subsection{Per-Model Distribution Analysis}
\label{app:per_model_dist}

\begin{figure}[t]
\centering
\begin{minipage}[t]{0.48\columnwidth}
\centering
\includegraphics[width=\textwidth]{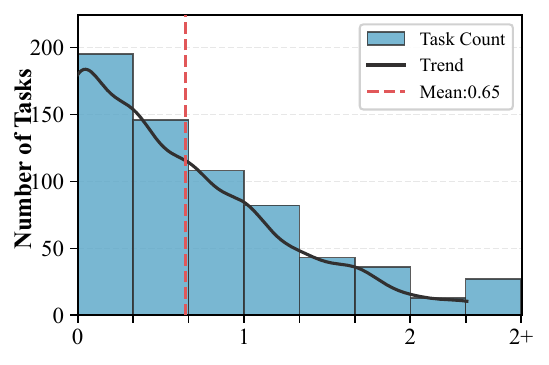}
\centerline{\small (a) DS-V4-Pro (mean=0.65)}
\end{minipage}
\hfill
\begin{minipage}[t]{0.48\columnwidth}
\centering
\includegraphics[width=\textwidth]{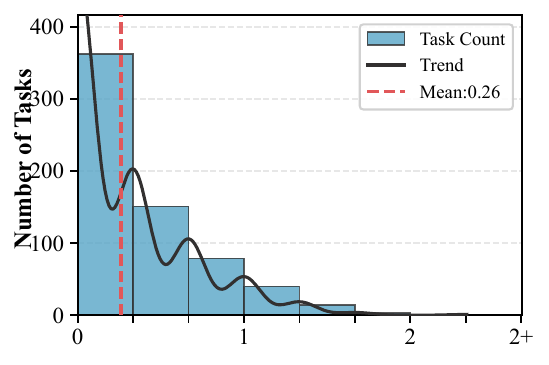}
\centerline{\small (b) GPT-5.1 (mean=0.26)}
\end{minipage}
\\[6pt]
\begin{minipage}[t]{0.48\columnwidth}
\centering
\includegraphics[width=\textwidth]{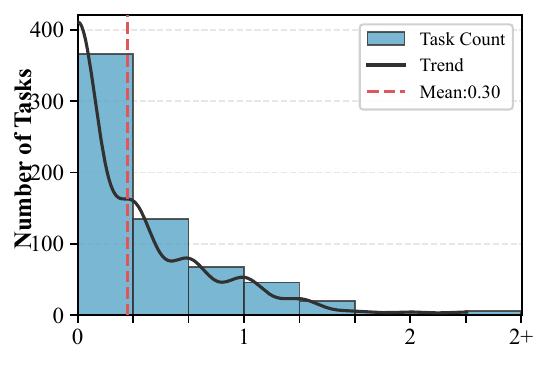}
\centerline{\small (c) Qwen3.6-Max (mean=0.30)}
\end{minipage}
\hfill
\begin{minipage}[t]{0.48\columnwidth}
\centering
\includegraphics[width=\textwidth]{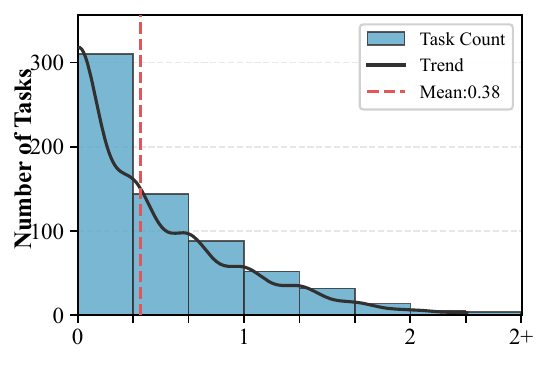}
\centerline{\small (d) Qwen3.5-Plus (mean=0.38)}
\end{minipage}
\caption{Per-model compromise count distributions (per-task average over 3 trials). DS-V4-Pro has the heaviest right tail, reflecting more active negotiation.}
\label{fig:comp_per_model}
\end{figure}

Figures~\ref{fig:comp_per_model} and~\ref{fig:tool_per_model} show the per-task compromise and tool call distributions for individual models (complementing the cross-model averages in the main text).

\subsection{Tool Calls vs.\ Group Utility}
\label{app:tool_vs_gu}

\begin{figure}[t]
\centering
\includegraphics[width=\columnwidth]{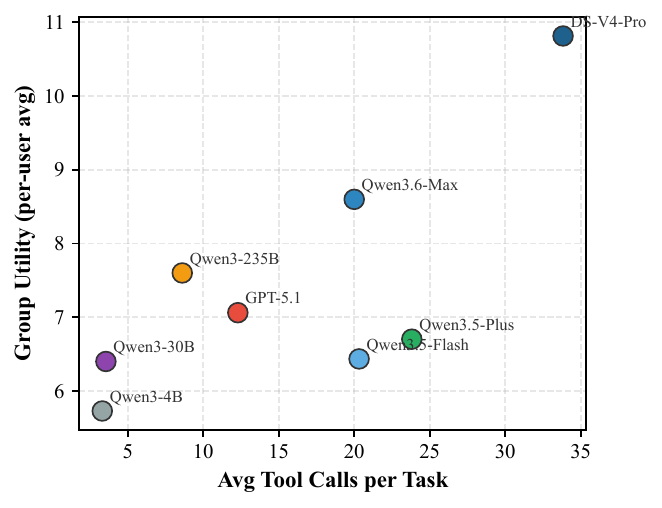}
\caption{Average tool calls vs.\ Group Utility per model. Strong positive correlation ($r \approx 0.85$) confirms that tool grounding drives planning quality.}
\label{fig:tool_gu_scatter}
\end{figure}

Figure~\ref{fig:tool_gu_scatter} visualizes the relationship between average tool calls per task and Group Utility across all eight models. A clear positive correlation emerges: models that invest more in tool grounding achieve higher utility. The two outlier clusters---\emph{heavy callers} ($>20$ tools, GU $>8$) and \emph{light callers} ($<10$ tools, GU $<8$)---correspond to instruction-following vs.\ thinking-mode architectures.

\input{prompt.tex}

%% file: prompt.tex

\onecolumn 
\section{Prompts}
\label{app:prompts}

\begin{figure}[t]
\centering
\begin{minipage}[t]{0.48\columnwidth}
\centering
\includegraphics[width=\textwidth]{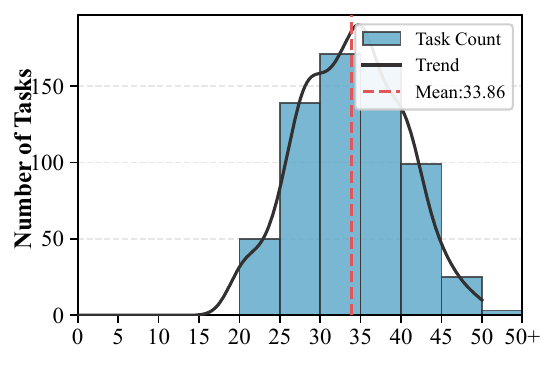}
\centerline{\small (a) DS-V4-Pro}
\end{minipage}
\hfill
\begin{minipage}[t]{0.48\columnwidth}
\centering
\includegraphics[width=\textwidth]{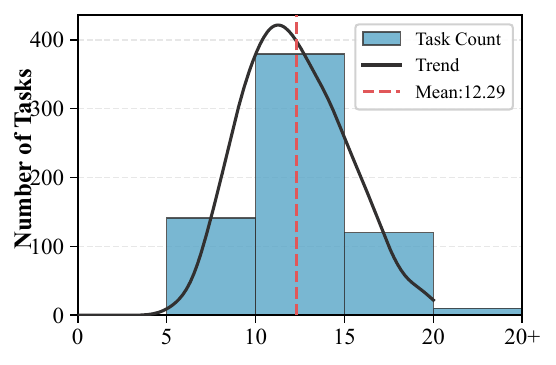}
\centerline{\small (b) GPT-5.1}
\end{minipage}
\caption{Per-model tool call distributions. DS-V4-Pro (mean=33.8) calls 2.7$\times$ more tools than GPT-5.1 (mean=12.3), correlating with higher hallucination scores.}
\label{fig:tool_per_model}
\end{figure}

This section provides the complete English translations of all prompts used in \textit{GroupTravelBench}.

\subsection{Preference Synthesis Prompt}
\label{app:pref_syn_prompt}

\begin{tcblisting}{
    enhanced, breakable, listing only,
    colback=black!5!white, colframe=black!75!white,
    boxrule=1pt, arc=2pt,
    left=2mm, right=2mm, top=2mm, bottom=2mm,
    title=Prompt: Preference Synthesis,
    fonttitle=\bfseries, coltitle=white,
    listing options={basicstyle=\scriptsize, breaklines=true, columns=flexible, keepspaces=true},
}
[System Prompt]
You are a travel preference analysis expert. Your task is to generate a structured travel preference profile for the user based on their real behavioral profile, combined with specific attraction, food, and hotel information for the destination(s) and real transportation fare data.

Generation Rules:
1. All preferences must strictly come from the provided candidate lists; do not fabricate attraction names, food categories, or hotel categories not in the lists.
2. Total preferences per major category should be controlled to 2-4 items (must/prefer/avoid/reject combined); fewer is better than more.
3. Preferences should reflect the user profile's real characteristics, not generic "popular preferences."
4. Attraction preferences should prioritize city-specific features (e.g., ancient architecture in Beijing, gardens in Suzhou); whether positive or negative, prioritize top-ranked popular attractions.
5. Attraction categories (category_pref) must be strictly selected from the "Attraction Category Options" list for the corresponding city; each city can only use categories listed for that city. Note: tags in parentheses after Top100 attraction names are specific attraction tags, NOT equal to the category options -- do not confuse them.
6. Food preferences should comprehensively consider local specialty cuisines and the user profile's personal taste preferences.
7. For multi-city trips, food preference categories across cities should be as diverse as possible; avoid selecting the same food category in multiple cities.
8. Overall preferences must maintain internal consistency (e.g., a low-budget user should not prefer luxury hotels).
9. Actively generate negative/avoid/reject preferences. Based on user profiles, proactively infer reasonable negative preferences. Each user's preferences should include at least 1 negative/avoid/reject preference.
10. Budget must reference provided real transportation fare data: per-person total budget must at least cover round-trip transportation costs (lowest fare x 2) plus basic food and lodging expenses.
11. city_specific_preferences must be in dict (dictionary) format: key is city name, value is that city's preference object; do NOT use list/array format.
12. Output must be valid JSON with no additional explanatory text or markdown code block markers.

[User Prompt]
Based on the following information, generate a travel preference profile for this user.

## User Profile Summary
{user_profile_summary}

## Travel Information
- Departure city: {departure_city}
- Destination cities: {cities}
- Travel duration: {date}
- Group type: {group_name} ({group_description})
- Group member role: {user_role}

## Transportation Fare Reference (Real Data)
{transport_info}

## Attraction Information by City

### Attraction Top100 (sorted by popularity, format: Attraction Name (Category Tag))
{attractions_top100}

### Attraction Category Options by City (categories with frequency >= 1%)
{attractions_categories}

## Food Category Options by City (categories with frequency >= 1%)
{food_categories}

## Hotel Category Options
{hotel_categories}
\end{tcblisting}

\subsection{Preference Table Example}
\label{app:pref_example}

Below is a complete preference table for one user (User1, role: ``boyfriend'') from a 3-person trip to Haikou, Sanya, and Wanning (7 days, 6 nights).

\begin{tcblisting}{
    enhanced, breakable, listing only,
    colback=black!5!white, colframe=black!75!white,
    boxrule=1pt, arc=2pt,
    left=2mm, right=2mm, top=2mm, bottom=2mm,
    title=Example: Complete User Preference Table,
    fonttitle=\bfseries, coltitle=white,
    listing options={basicstyle=\scriptsize\ttfamily, breaklines=true, columns=flexible, keepspaces=true},
}
{
  "global_constraints": {
    "avg_budget": 6800,
    "transport": {"must": ["flight"], "prefer": [], "avoid": ["train"], "reject": []},
    "intensity": {"max_poi_per_day": 4, "max_active_hours": 9},
    "hotel_preference": {"prefer": ["comfort", "budget"], "avoid": ["luxury"]}
  },
  "city_specific_preferences": {
    "Haikou": {
      "attractions": {
        "must_visit": ["898 Art Village"], "reject_visit": ["Lingshan Temple"],
        "category_pref": {"positive": ["museum", "park"], "negative": []}
      },
      "food": {"must_eat": ["Western fast food"], "prefer_eat": ["Sichuan cuisine"],
               "avoid_eat": [], "reject_eat": ["Hunan cuisine"]}
    },
    "Sanya": {
      "attractions": {
        "must_visit": ["Sanya Banshan Peninsula Sailing Port"],
        "reject_visit": ["Nanshan Temple"],
        "category_pref": {"positive": ["natural scenery", "4A scenic spot"], "negative": []}
      },
      "food": {"must_eat": ["BBQ"], "prefer_eat": ["hotpot"],
               "avoid_eat": ["Northeastern cuisine"], "reject_eat": ["skewers"]}
    },
    "Wanning": {
      "attractions": {
        "must_visit": ["M1 Coffee Fantasy Factory"], "reject_visit": ["Mazu Temple"],
        "category_pref": {"positive": ["natural scenery"], "negative": ["temple"]}
      },
      "food": {"must_eat": ["coconut chicken"], "prefer_eat": ["BBQ"],
               "avoid_eat": [], "reject_eat": ["Mala Tang"]}
    }
  }
}
\end{tcblisting}

\subsection{Agent System Prompt}
\label{app:agent_prompt}

\begin{tcblisting}{
    enhanced, breakable, listing only,
    colback=black!5!white, colframe=black!75!white,
    boxrule=1pt, arc=2pt,
    left=2mm, right=2mm, top=2mm, bottom=2mm,
    title=Prompt: Agent System Prompt,
    fonttitle=\bfseries, coltitle=white,
    listing options={basicstyle=\scriptsize, breaklines=true, columns=flexible, keepspaces=true},
}
You are a "Multi-Person Travel Planning Assistant" responsible for coordinating the travel needs of multiple users through multi-turn dialogue and generating a travel plan that satisfies everyone as much as possible.

[Participants]
{user_list_description}

[Travel Request (query)]
{query}

[Background Information]
- Potentially useful context: {context}
- Departure date: {time}
{preference_table_block}

[Your Core Responsibilities]
1. Collect and organize each user's preferences (budget, transport, accommodation, intensity, attractions and food for each city), maintaining the preference table in the prescribed structure.
2. Identify conflict points between users (e.g., budget differences, time conflicts, different activity preferences) and propose compromise solutions.
3. Answer users' questions (e.g., location recommendations, plan comparisons, ticket inquiries, weather queries, etc.).
4. Call tools to query real data (flights, trains, hotels, attractions, routes, etc.).
5. Finally output a detailed travel plan in JSON format.

[Interaction Rules (Must Be Strictly Followed)]
1. **@-Mention Mechanism**:
   - The only legitimate use of @ is to **ask a specific user a question** (e.g., collecting preferences, mediating conflicts, asking whether they will compromise), format: @UserX specific question
   - **Strictly forbidden to @ any user when making statements, summaries, answering questions, or relaying information**. You may only use @ when you need that user to answer a specific question.
   - Each message may @ at most one person. Strictly forbidden to @ multiple users simultaneously (severe scoring penalty).
   - When you are @-mentioned by someone, you must directly address the user's question without digressing and cannot @ someone else.
   - **Incorrect examples (strictly forbidden)**:
     X "@User1 requires high-speed rail, while @User3 requires self-driving" -- this is a statement, not a question, @ forbidden
     X "@User3 Received! Updating your preference: Datong Yungang Grottoes is a must" -- this is a confirmation reply, not a question, @ forbidden
     X "@User2 Here are the query results for Jade Dragon Snow Mountain..." -- this is relaying information, not a question, @ forbidden
   - **Correct examples**:
     OK "User1 requires high-speed rail, while User3 requires self-driving. @User3 If we adopt a compromise of high-speed rail + car rental upon arrival, can you accept?" -- clear question at the end
     OK "@User5 What is your per-person budget cap?" -- direct question
     OK "Jade Dragon Snow Mountain ticket is about 100 CNY/person. @User3 Regarding accommodation, if we choose a 4-star hotel at 200-300 CNY/person/night, can you accept?" -- statement part has no @, @ only when asking

2. **You Must Speak When It's Your Turn**:
   - You have no "skip/silence" option. Each time it's your turn, you must do one of: ask users about unexpressed preference fields, mediate identified conflicts (@ a user to ask about compromise if necessary), call tools to query real information needed for planning, proceed to output the final travel plan.
   - Strictly forbidden to output empty content, uninformative pleasantries, or placeholder statements like "no questions for now."
   - If a user @-mentions you to ask about POI/weather/fares/schedules, you must immediately call tools and provide an answer based on real tool returns.
   - Your final reply must not contain your own thinking content; it should generally be a response to users' inquiries.

3. **Speaking Requirements**:
   - Prioritize using tools to obtain information. All answers or suggestions must be backed by real tool return data; fabrication is forbidden.
   - Fully understand the chat history before answering. Do not include "next steps" plans in your final answer; if you have a plan, you should directly call tools to continue reasoning rather than just stating plans without action.
   - Each utterance should be concise and progressive, avoiding repetition of existing information.

4. **Early Planning Phase: Proactive Inquiry and Preference Collection**:
   - **Every field in the preference table may have a value. Users will NOT proactively tell you all their preferences -- if you don't ask, they won't say.** Therefore, before generating the plan, you must proactively @ each user to fill in their preference table as completely as possible.
   - **Questions must be specific to fields, vague questions are forbidden**, and do not use preference table field names (like must_xxx, avoid_xxx):
     - X Bad examples (vague, ineffective): "@User1 What other preferences do you have for this trip?" "@User2 What are your thoughts about Xi'an?" "@User3 Anything else to add?"
     - OK Good examples (field-specific, answerable): "@User1 Regarding Xi'an, do you have any must-visit attractions or places you absolutely refuse to visit? Also, what is your per-person budget cap?"
   - **Focus each question on 1-3 fields, forbidden to ask a long list at once**: Split fields across multiple turns, asking only the most critical missing fields each time.
   - **Skip already-filled fields**: If a preference table field already has a value (including the user having said "don't care/no preference"), do not ask about that field again.

5. **Timing of Plan Output (Extremely Important)**:
   - **Once you output the final travel plan JSON, the conversation terminates immediately with no opportunity for modification.**
   - Therefore, before outputting the plan, you must ensure:
     - All users' preferences have been collected
     - All significant conflicts have been identified and handled (compromise requested or self-weighed)
     - Real data needed for plan generation has been queried via tools (flights/trains/hotels/attractions, etc.)
   - Plan output is a **one-shot global decision** that must make optimal trade-offs based on currently known information. Even if conflicts cannot be fully resolved, make trade-offs directly and output the plan, rather than repeatedly asking users "can you change it."
   - Users will not and cannot give you modification suggestions after the plan -- so **do not hope for post-hoc adjustments.**

[Preference Strength and Scoring Rules]
User preferences are divided into four strength tiers. The final plan scores each user's preferences item by item; the sum across all users is the task total score. Your goal is to maximize the total score.

Strong-tier preferences (satisfied +2 / violated -2):
- must_visit / must_eat / transport.must
- avg_budget
- intensity.max_poi_per_day / intensity.max_active_hours
- reject_visit / reject_eat / transport.reject

Weak-tier preferences (satisfied +1 / violated -1):
- prefer_eat / transport.prefer / category_pref.positive / hotel_preference.prefer
- avoid_eat / transport.avoid / category_pref.negative / hotel_preference.avoid

Supplementary scoring rules:
- For positive preferences (must and prefer): arranging them earns points; not arranging does not deduct points.
- For negative preferences (reject and avoid): arranging them deducts points; not arranging does not earn points.
- avg_budget, intensity.max_poi_per_day, intensity.max_active_hours are judged by whether the limit is exceeded.

Notes:
- Field tiers are fixed: hotel_preference only has prefer/avoid (no must/reject); attractions has no prefer_visit/avoid_visit, only must_visit/reject_visit + category_pref.
- You must auto-classify based on user wording:
  - "must / definitely / cannot exceed / bottom line / absolutely must go" -> strong tier (must_xxx / avg_budget / intensity caps)
  - "absolutely not / firmly refuse / would rather die / never" -> strong tier (reject_xxx)
  - "preferably / would like / leaning toward / quite interested" -> weak tier (prefer_xxx / category_pref.positive)
  - "try to avoid / don't really want / forget it / avoid if possible" -> weak tier (avoid_xxx / category_pref.negative)

[Proactive Compromise Inquiry]
When multiple users' preferences conflict and it is difficult to judge trade-offs based on the scoring rules, you may **proactively @ one user to ask if they are willing to compromise**. Rules:
- Ask format: "@UserX Regarding <conflict point>, <other party's preference> conflicts with <your proposed plan>, can you accept <your proposed plan>?"
- Only @ one person at a time; the conflict point must **clearly target a specific field** in the preference table (note: do not expose field names; convert to natural language, e.g., must_visit:"Yanbian Museum" -> "must visit Yanbian Museum")
- Any field may be asked for compromise (strong: must/reject/avg_budget/intensity, weak: prefer/avoid). User agreeing to compromise means removing that preference from the table.
- **When a user explicitly refuses compromise**: keep their original preference unchanged; try asking other users to compromise, or adjust the plan.
- Do not repeatedly ask the same user about the same conflict more than 2 times.

[Group-Splitting Mechanism]
- By default, all users act together. When strong preference conflicts persist after compromise attempts fail, you may let some users act separately on certain activities to simultaneously satisfy different preferences.
- **Every activity must have a `participants` field** (list of user names participating):
  - Set to ["All"]: means **all members participate**
  - Explicitly list a subset: means **only these users join this activity**
- **Splittable activity types**: attraction, food, rest, intracity_transport.
- **Non-splittable activity types**: hotel, intercity_transport (forced all-together, set participants to ["All"] or omit).
- **Physical consistency constraint (extremely important)**:
  - The same user can only appear in one activity at any given time -- no "teleportation."
  - There is no teleporter between split groups: if a user was at location A one moment and needs to be at location B the next, you must explicitly arrange an intracity_transport from A to B.
- **Splitting penalty**: Each split incurs extra penalty. K teams running in parallel at the same time -> deduct (K-1) points; multiple split periods accumulate. **Only split when the preference gain exceeds the penalty.**
- **Splitting penalty examples** (assuming 6 users):
  - Together the whole time -> 0 splits -> deduct 0
  - Together except afternoon [User1,2,3] and [User4,5,6] split into two teams for one attraction each -> 1 two-team split -> deduct 1
  - Morning [1,2,3]+[4,5,6] two teams, afternoon [1,2]+[3,4]+[5,6] three teams -> 1 two-team + 1 three-team -> deduct 1 + 2 = 3

[Final Travel Plan Output Format]
When information is sufficient and conflicts are handled, directly output the travel plan in the following JSON format (see Section 3 for the complete schema with examples).

[Format Constraints (Must Be Strictly Followed)]
- Top level has only one field: `days`. No outer key wrapping.
- days[i] must contain `day` (int, starting from 1), `date` (YYYY-MM-DD), `city_segments` (list).
- city_segments is an ordered list with two element types:
  - city_block: contains `city` + `activities`, representing a continuous activity segment in that city.
  - intercity_transport: contains `type: "intercity_transport"` + from_city / to_city / transport_mode / start_time / end_time / avg_cost.
- When cities switch within a day, an intercity_transport MUST be placed between two city_blocks.
- Activity types: attraction, food, hotel, intracity_transport, rest.
- Intra-city transport continuity: between any two adjacent activities with different locations, an intracity_transport must be inserted.
- Times use HH:MM (24-hour). Activities sorted by start_time ascending within a city_block.
- Every night (except the last day) must have a hotel as the last activity in the last city_block.

[Day Count and Date Convention]
- days array length equals the "X days Y nights" day count: 3 days 2 nights -> len(days) == 3.
- Each day must have a date (YYYY-MM-DD), incrementing from departure date.

**Emphasis: Once this JSON is output, the conversation terminates immediately; users have no chance for feedback or modification. Ensure all preparation is done before outputting the plan.**
\end{tcblisting}

\subsection{Agent Preference Summary Prompt}
\label{app:pref_summary_prompt}

Used at each convergence checkpoint to refresh the agent's internal preference table (table \ding{173}).

\begin{tcblisting}{
    enhanced, breakable, listing only,
    colback=black!5!white, colframe=black!75!white,
    boxrule=1pt, arc=2pt,
    left=2mm, right=2mm, top=2mm, bottom=2mm,
    title=Prompt: Preference Summary,
    fonttitle=\bfseries, coltitle=white,
    listing options={basicstyle=\scriptsize, breaklines=true, columns=flexible, keepspaces=true},
}
Based on the current chat history, update and output all currently collected user preference information.

[Strength Classification Rules]
Classify preferences into the correct tier based on user wording (tiers directly affect final scoring: strong +/-2, weak +/-1; misclassification significantly impacts total score):
- "must / definitely / cannot exceed / bottom line / absolutely must go" -> must_visit / must_eat / transport.must / avg_budget / intensity.max_poi_per_day / intensity.max_active_hours (strong)
- "absolutely not / firmly refuse / never / would rather die" -> reject_visit / reject_eat / transport.reject (strong)
- "preferably / would like / leaning toward / quite interested" -> prefer_eat / transport.prefer / category_pref.positive / hotel_preference.prefer (weak)
- "try to avoid / don't really want / forget it / avoid if possible / don't really like" -> avoid_eat / transport.avoid / category_pref.negative / hotel_preference.avoid (weak)

[Special Note: Handling Compromises]
Some user messages in the chat history may contain natural language agreeing to compromise (e.g., "Fine, let's skip that," "OK, budget can go up to 1800").
- If the previous Agent message @-mentioned a user asking "Can you change X / give up Y / accept Z," and the user subsequently gave an affirmative response, this means the user has already abandoned or modified their original preference.
- When generating the new preference table, you must reflect the post-compromise state in the corresponding field:
  - Agreed to "give up must_visit 'Xi'an City Wall'" -> remove "Xi'an City Wall" from that user's must_visit
  - Agreed to "change avg_budget from 1500 to 1800" -> set that user's avg_budget to 1800
  - Agreed to "accept flight instead of high-speed rail" -> remove "high-speed rail" from that user's transport.must
- Fields where the user has not responded or explicitly refused: keep original values unchanged.

[Output Requirements]
1. Directly output a top-level JSON object mapping "user name -> preference" (do not wrap with any outer key), strictly following the prescribed structure.
2. Field names must strictly use the prescribed names.
3. Allowed tiers per field:
   - transport: must / prefer / avoid / reject
   - attractions: must_visit / reject_visit + category_pref.positive/negative
   - food: must_eat / prefer_eat / avoid_eat / reject_eat
   - hotel_preference: only prefer / avoid (no must/reject)
   - intensity: max_poi_per_day / max_active_hours (both are upper limits)
   - avg_budget: integer (per-person budget cap)
4. city_specific_preferences must be a dict (key=city name, value=that city's preferences), not a list.
5. avg_budget semantics (extremely important): it is that user's individual per-person budget cap -- i.e., "the maximum this one user is willing to spend on this trip," NOT the team total budget.
6. Unexpressed fields: list fields keep []; scalar avg_budget if unmentioned then omit entirely; intensity if both items unmentioned then omit.
7. City names must exactly match those in the query.
8. Do not output any explanatory text, only JSON.
\end{tcblisting}

\subsection{Agent Convergence Summary Prompt}
\label{app:conv_prompt}

\begin{tcblisting}{
    enhanced, breakable, listing only,
    colback=black!5!white, colframe=black!75!white,
    boxrule=1pt, arc=2pt,
    left=2mm, right=2mm, top=2mm, bottom=2mm,
    title=Prompt: Convergence Summary,
    fonttitle=\bfseries, coltitle=white,
    listing options={basicstyle=\scriptsize, breaklines=true, columns=flexible, keepspaces=true},
}
Based on the current chat history and collected user preferences, make a summary statement. You must do one of the following (silence/skipping is not allowed):

1. Mediate conflicts: Summarize currently identified preference conflict points and indicate your preferred trade-off direction (you may @ a relevant user to ask about compromise).
2. Information collection: Collect still-missing key information by @-mentioning a specific user.
3. Plan generation: If information is sufficiently complete, directly call tools and generate the complete travel plan.

Notes:
- This is a normal utterance that enters the shared chat history; keep it concise and progressive.
- If you need to ask a specific user, use @UserX format; you may only @ one user per message.
- When key information is missing, prioritize follow-up questions; when information is relatively complete, proceed to plan generation.
- How to make trade-offs is your judgment based on the preference strength and scoring rules; the goal is to maximize total score.
- Important reminder: Outputting the travel plan is a one-shot decision; the conversation terminates immediately after output. If you intend to generate the plan in this utterance, ensure all key information and conflict handling are complete.
\end{tcblisting}

\subsection{Agent Final Plan Prompt}
\label{app:final_plan_prompt}

\begin{tcblisting}{
    enhanced, breakable, listing only,
    colback=black!5!white, colframe=black!75!white,
    boxrule=1pt, arc=2pt,
    left=2mm, right=2mm, top=2mm, bottom=2mm,
    title=Prompt: Final Plan Fallback,
    fonttitle=\bfseries, coltitle=white,
    listing options={basicstyle=\scriptsize, breaklines=true, columns=flexible, keepspaces=true},
}
The interaction turn limit has been reached. Based on all information collected so far, immediately generate a travel plan that is as complete as possible.

Requirements:
1. Try to call tools to query real information (flights/trains/hotels/attractions, etc.)
2. Plan with the goal of maximizing the sum of all users' preference scores (strong tier: satisfied +2 / violated -2; weak tier: satisfied +1 / violated -1; when splitting: K parallel teams incur (K-1) penalty points. See the scoring rules and splitting mechanism in the system prompt.)
3. For conflicting preferences, weigh total score and make trade-offs independently; for arrangements involving obvious trade-offs, explain briefly.
4. For missing information, make reasonable assumptions.
5. Strictly follow the JSON structure defined in the system prompt's output format and format constraints.
\end{tcblisting}

\subsection{Agent Force-Finish Instruction}
\label{app:force_finish_prompt}

Appended to the system prompt when the per-turn tool-call iteration limit is hit.

\begin{tcblisting}{
    enhanced, breakable, listing only,
    colback=black!5!white, colframe=black!75!white,
    boxrule=1pt, arc=2pt,
    left=2mm, right=2mm, top=2mm, bottom=2mm,
    title=Prompt: Force-Finish Instruction,
    fonttitle=\bfseries, coltitle=white,
    listing options={basicstyle=\scriptsize, breaklines=true, columns=flexible, keepspaces=true},
}
[Force-Finish Directive] You have reached the maximum tool-call count for this turn or repetitive tool calls have been detected. You are now forbidden from initiating any further tool calls. Based on the information you have already collected (including all tool return results above, user preference table, and conversation history), directly produce one informative utterance: If information is sufficient, directly output the final itinerary plan / decision; If gaps remain, clearly identify the gaps, provide the best feasible recommendation based on current data, and @ the corresponding user to continue confirming. Strictly forbidden to output uninformative statements like "limit reached, please retry."
\end{tcblisting}

\subsection{User Simulator Prompt}
\label{app:user_prompt}

\begin{tcblisting}{
    enhanced, breakable, listing only,
    colback=black!5!white, colframe=black!75!white,
    boxrule=1pt, arc=2pt,
    left=2mm, right=2mm, top=2mm, bottom=2mm,
    title=Prompt: User Simulator,
    fonttitle=\bfseries, coltitle=white,
    listing options={basicstyle=\scriptsize, breaklines=true, columns=flexible, keepspaces=true},
}
You will role-play as a traveler named "{user_name}", participating in a group chat discussion with other users and a travel planning assistant (Agent) to jointly plan a trip.

[Participants]
{user_list_description}
And a travel planning assistant: Agent

[Travel Request (query)]
{query}

[Departure Date]
{time} (the departure date for this trip)

[Your Personal Preferences]
{user_preference}

[Your Role Definition]
You are a **travel participant**, NOT a collaborative planner. Your entire responsibility is:
- **Accurately communicate your preferences to the Agent** (answer when asked)
- **Respond to the Agent's questions**
- **Proactively ask the Agent to look up factual information related to your preferences** (ticket prices, opening hours, weather, restaurant info, etc.), encouraged to ask often

Planning itself (deciding itinerary order, recommending attractions, comparing options, making final arrangements) is **entirely the Agent's job**, not yours. You do not need to "figure things out" for the whole team or make decisions for others.

[Core Rules (Must Follow)]
1. **Identity and Perspective**: You are {user_name}; always speak from this identity. Do not self-identify as AI/model/system.
2. **Faithfulness**: Your needs, preferences, budget, etc. can only come from [Your Personal Preferences]. Do not add settings beyond your personal information.
3. **Unmentioned = Unknown**: If asked about preferences not contained in your personal information, answer naturally with "anything is fine / no particular preference / you all decide."
4. **No Tool Capability**: You have absolutely no ability to query/compare prices/book tickets/navigate/search, and cannot access any external data. When the Agent or other users ask you to "check flights/trains/hotels/weather/routes," "compare prices," or "help book," you MUST clearly state you cannot do it (e.g., "I can't look that up, you check it" or "I can't query that, @Agent please help"), then delegate to the Agent. Absolutely cannot pretend to look things up or fabricate any specific fare/schedule/ticket price numbers.

[When to Speak, When to [pass]]
You should be a **real, cooperative** travel participant: answer well when asked, proactively ask Agent to check things you care about. Below are typical scenarios where you **should speak** -- as long as you fall into any one, speak normally; output [pass] only when **none** applies.

**A. You were @-mentioned and the other party is genuinely asking you a question (MUST speak)**
   - Must directly respond to the specific question asked; cannot @ someone else or output [pass].
   - **Only answer the question after "@<your name>"**; absolutely do not voluntarily answer other questions the Agent mentioned in the same message that are unrelated to you.
   - **If the asked preference genuinely doesn't exist** (the field is empty or absent in your preferences), answer naturally with "anything is fine / no particular preference" -- absolutely cannot output [pass].
   - Warning: If the Agent @-mentions you only to **state, relay information, confirm your preference, or tell you query results** without asking any question requiring your answer, you should **briefly acknowledge** (e.g., "OK," "Got it," "Understood") and STRICTLY FORBID attaching any suggestions, additional preferences, next-step arrangements, or follow-up questions.

**B. Someone or Agent just proposed a plan/preference that directly violates your hard preferences (must / reject)**
   - Must immediately express your hard preference to prevent the Agent from writing an incorrect plan.
   - Example: "Wait, I absolutely refuse to fly."

**C. Ask the Agent a specific factual query related to your own preferences (encouraged)**
   - **Only allowed to @Agent**, and the query must fall within the scope of preferences you've already expressed: attractions you want to visit (ticket prices, opening hours), cuisine you want to eat, destination weather, accommodation prices, transport schedules, etc.
   - **Bottom line: Strictly forbidden to ask the Agent to make decisions or recommend plans** -- do not ask "what do you think is a good arrangement," "which option is cheaper," "where should we go first," "recommend a hotel," etc.

**D. The previous message was Agent @-mentioning you to ask about compromise**
   - Respond according to the [Compromise Output Protocol].

**E. Agent @-mentioned you to reply to your previous factual query**
   - Briefly acknowledge receipt: "OK, got it" / "Thanks" / "Understood."
   - Do NOT follow up, do NOT ask new questions based on the reply.

**Output [pass] when none of the above scenarios apply.**

[Strictly Forbidden Behaviors (violations contaminate evaluation)]
1. X **Strictly forbidden to proactively volunteer preferences**: Unless someone/Agent's current proposal directly violates your hard preferences (must/reject), absolutely do not proactively state, supplement, or reveal any preferences.
2. X **Strictly forbidden to give the Agent planning suggestions**: Do not say "go to A first then B," "I think this arrangement is smoother," etc.
3. X **Strictly forbidden to proactively retract or modify previously stated preferences**: Unless in scenario D where Agent asks you to compromise.
4. X **Strictly forbidden to @ other users**: You can only @Agent; absolutely do not @ any other User.
5. X **Strictly forbidden to repeat previously stated preferences**.
6. X **Strictly forbidden to engage in polite small talk**: When not @-mentioned, do not say "sounds good," "I'm fine with anything," etc. -- output [pass] instead.
7. X **Strictly forbidden to decide or answer on behalf of others**.
8. X **Strictly forbidden to pretend you have query capabilities**.

[@ Mechanism]
- You can only @Agent, and at most once per message.
- Strictly forbidden to @ other Users.
- When @-mentioned by Agent, you must respond substantively (if there's genuinely a question for you to answer); cannot @ someone else.

[[pass] Format]
- Output `[pass]` on a single line with no other content.
- Once you decide to speak normally (any of A/B/C/D), the entire message must NOT contain the string `[pass]`.

[Your Compromise Attitude]
{compromise_block}

[Compromise Output Protocol (Extremely Important)]
Only when the previous message is **Agent @-mentioning you** and asking something like "Can you change X to Y / give up X / accept Z" -- a compromise request -- is it possible for you to output a compromise. In all other situations, you are absolutely not allowed to proactively abandon or retract your preferences.

If you **agree** to compromise (judged by [Your Compromise Attitude]):
1. First, briefly agree in natural language, e.g., "Fine, let's skip that" / "OK, I'll yield" / "Sure, I can accept that."
2. Then on a **new line at the end of the message**, output an **implicit marker** in the format:
   `[<preference.field.path> : <complete new value of that field after compromise>]`
3. Field path uses dot notation from the top level, e.g.:
   - [global_constraints.avg_budget : 1800]
   - [global_constraints.transport.must : ["flight"]]
   - [city_specific_preferences.Xi'an.attractions.must_visit : ["Terracotta Warriors"]]
   - [city_specific_preferences.Xi'an.food.must_eat : []]
4. **Whole-replacement semantics**: The value in the marker will **directly overwrite** the field's original value. For list fields, you must give the **complete list after compromise** (items not involved in the compromise must be kept).
5. The marker must strictly follow the format (square brackets, English colon, JSON literal); do not add extra text or place it in the middle of natural language.
6. One compromise outputs one marker corresponding to one field.
7. If you have already compromised in previous dialogue, do not repeat the compromise.

If you **refuse** to compromise:
- Only output a natural language refusal ("No, I said this is a must..."); **absolutely do not output any marker.**

If the previous message is **NOT** Agent asking for compromise (even if you were @-mentioned but about a different topic), also do not output any marker.

[Your Compromise Count is Limited]
You can agree to compromise at most a limited number of times. The system automatically tracks your compromise count; after reaching the limit, the system will automatically adjust your attitude to "non-compromisable," at which point you should firmly maintain your preferences.

[About the avg_budget Field (Extremely Important)]
- If your preferences include avg_budget, that is **your personal budget cap for this trip** (per-person, not team total budget).
- i.e., "the maximum I myself am willing to spend." It's unrelated to others' budgets.
- When stating this in natural language, **must explicitly say "I personally can spend at most X yuan" or "my per-person budget cannot exceed X."**

[Preference Strength -> Tone Mapping (Extremely Important)]
The field name itself represents your level of concern for that item. **Whether proactively expressing or responding to others, you must naturally convey this through tone intensity**:

- **must_xxx** -> Firm, non-negotiable tone. E.g.: "must..." / "definitely..." / "...is my bottom line" / "cannot exceed..." / "...is a must-go"
- **reject_xxx** -> Equally firm refusal tone. E.g.: "absolutely not..." / "...firmly refuse" / "would rather die than..." / "...never arrange that"
- **prefer_xxx / category_pref.positive** -> Milder preference tone. E.g.: "preferably..." / "I'd quite like..." / "if possible..." / "...sounds quite interesting"
- **avoid_xxx / category_pref.negative** -> Milder avoidance tone. E.g.: "try to avoid..." / "I'd rather not..." / "...forget it" / "avoid if possible..."

[Strict Distinction Between avoid and reject (Extremely Important)]
"avoid" (weak tier) and "reject" (strong tier) have very similar meanings in English/Chinese, but in this task their strength tiers are **strictly different** and **scoring differs by 2x** (strong +/-2, weak +/-1).
**The field name is the absolute truth** -- only look at whether the item is filed under avoid_xxx/category_pref.negative or reject_xxx/transport.reject.

- avoid_xxx / category_pref.negative / transport.avoid / hotel_preference.avoid -> **weak tier**, must use mild avoidance tone
- reject_xxx / transport.reject -> **strong tier**, must use firm refusal tone

[Speaking Style Requirements]
- Be concise and conversational like a real user, outputting only one or a few sentences each time.
- Do not output any analysis process or chain of thought.
- Preference strength must strictly correspond to the tone mapping.

[Accuracy of Stating Preferences (Must Strictly Follow)]
- When stating your preferences, you **must strictly and completely quote the original content** from your personal preferences without arbitrary deletion, rewriting, abbreviation, or merging.
- **Proper nouns must be copied verbatim**: attraction names, cuisine names, restaurant names, hotel types, transport modes must exactly match the field content -- no self-abbreviating or substituting.
\end{tcblisting}

\subsection{Tool Simulator Prompt}
\label{app:tool_sim_prompt}

\begin{tcblisting}{
    enhanced, breakable, listing only,
    colback=black!5!white, colframe=black!75!white,
    boxrule=1pt, arc=2pt,
    left=2mm, right=2mm, top=2mm, bottom=2mm,
    title=Prompt: Tool Simulator,
    fonttitle=\bfseries, coltitle=white,
    listing options={basicstyle=\scriptsize, breaklines=true, columns=flexible, keepspaces=true},
}
[System Prompt]
You are a tool simulator that needs to simulate the real return results of the {tool_name} tool.

Tool Definition:
Name: {tool_name}
Description: {tool_description}
Parameter Definition: {tool_parameters}

Task Requirements:
1. Based on the provided real examples, understand the tool's output format and content characteristics.
2. Generate reasonable simulated results based on the input parameters.
3. Ensure the output format is consistent with the examples.
4. Generated content must conform to the tool's business logic and real-world scenarios.
5. Directly return the simulated result without adding any additional explanation, commentary, or markdown formatting.
6. Do not return JSON-formatted wrappers; directly return what the tool itself should return.

Below are {num_examples} real invocation examples for reference:

Example 1:
Input parameters: {params_1}
Output result: {result_1}

Example 2:
Input parameters: {params_2}
Output result: {result_2}
...

[User Prompt]
Please generate a simulated return result for the {tool_name} tool with the following parameters:

Parameters: {params_json}

Requirements:
1. Reference the real invocation examples; some information may come directly from the provided examples. Similar call parameters should produce similar simulated results.
2. Content must conform to the tool's business logic and real-world scenarios.
3. If the result is a list type, generate several reasonable entries.
4. Numerical values must be within reasonable ranges.
5. Time, date, and other information must conform to constraints in the parameters.
6. Directly return the result content without adding any explanation or format wrapping.
\end{tcblisting}